\pdfoutput=1
\documentclass[11pt]{article}
\usepackage[]{acl}

\usepackage{amsmath,amsfonts,bm}

\def\eqref#1{equation~\ref{#1}}

\def\1{\bm{1}}

\DeclareMathAlphabet{\mathsfit}{\encodingdefault}{\sfdefault}{m}{sl}
\SetMathAlphabet{\mathsfit}{bold}{\encodingdefault}{\sfdefault}{bx}{n}

\usepackage{times}
\usepackage{latexsym}
\usepackage[T1]{fontenc}
\usepackage[utf8]{inputenc}
\usepackage{microtype}

\usepackage{multirow}
\usepackage{graphicx}
\usepackage{xcolor}
\usepackage{booktabs}
\usepackage{amsthm}
\usepackage{mathrsfs}
\usepackage{verbatim}

\usepackage{amsmath,amssymb}
\usepackage{algorithm}
\usepackage{algorithmic}

\title{
Sharpness-Aware Minimization Improves Language Model Generalization
}

\author{
Dara Bahri, Hossein Mobahi, Yi Tay\\
Google Research\\
\texttt{\{dbahri,hmobahi,yitay\}@google.com}\\
}
\date{}

\begin{document}

\maketitle

\begin{abstract}
The allure of superhuman-level capabilities has led to considerable interest in language models like GPT-3 and T5, wherein the research has, by and large, revolved around new model architectures, training tasks, and loss objectives, along with substantial engineering efforts to scale up model capacity and dataset size. Comparatively little work has been done to improve the generalization of these models through better optimization. In this work, we show that Sharpness-Aware Minimization (SAM), a recently proposed optimization procedure that encourages convergence to flatter minima, can substantially improve the generalization of language models without much computational overhead. We show that SAM is able to boost performance on SuperGLUE, GLUE, Web Questions, Natural Questions, Trivia QA, and TyDiQA, with particularly large gains when training data for these tasks is limited.
\end{abstract}

\section{Introduction}
Over the last several years, remarkable progress has been made within the domain of natural language understanding, with machine-learned models able to solve some tasks at near or above human-level performance. This progress has, by and large, been fueled by research centered around 1) better \emph{inductive biases}, such as the attention-enabled Transformer architecture~\citep{vaswani2017attention}, 2) the clever leverage of massive corpora of textual data that was historically disregarded as ``unlabeled,'' usually in the form of pre-training objectives that strive to teach the model the structure of language~\citep{radford2019language,devlin2018bert}, 3) scaling up model capacity and the methods to support it~\citep{shazeer2018adafactor}, 4) multi-task learning~\citep{raffel2019exploring}, and lastly, 5) larger and more diverse datasets along with ever-improving benchmarks that attempt to test the true capabilities of these models. Although these efforts all share the single goal of improving the model's generalization, doing so by explicit changes to the optimization of the loss function has received less attention in comparison.

Recently, motivated by the both empirical and theoretical findings that flatter minima lead to better generalization~\citep{kleinberg2018alternative,keshar,chaudhari2019entropy,smith2017bayesian}, \citet{foret2020sharpness} proposed a novel modification to vanilla stochastic gradient descent they term ``Sharpness-Aware Minimization,'' or SAM. They show theoretically and empirically that optimizing with SAM encourages convergence to flatter points in the loss landscape and with it comes the anticipated improvement in out-of-sample error. While their empirical findings are limited to computer vision tasks and datasets using convolutional neural networks (ResNets), follow-up work~\citep{chen2021vision} showed how SAM is particularly effective on Vision transformers (ViTs)~\citep{dosovitskiy2020image} and MLP-Mixers~\citep{tolstikhin2021mlp}, architectures that are more prone than convolutional ones to land in sharp minima. Crucially, they show that when equipped with SAM, ViTs outperform ResNets of similar size and throughput \emph{without the need for large-scale pre-training}.

Encouraged by wins in the vision domain, we ask whether SAM can deliver similar gains in the language domain. Our contributions are as follows:
\begin{enumerate}
\item We show that blithely applying SAM when fine-tuning public pre-trained checkpoints of the text-to-text transformer (T5)~\citep{raffel2019exploring} and its multilingual counterpart, mT5~\citep{xue2020mt5} on SuperGLUE~\citep{wang2019superglue}, GLUE~\citep{wang2018glue}, TyDiQA-GoldP~\citep{clark2020tydi} and the Closed-Book Question Answering (CBQA) tasks from \citet{roberts2020much} -- Web Questions~\citep{berant2013semantic}, Natural Questions~\citep{kwiatkowski2019natural}, and Trivia QA~\citep{joshi2017triviaqa} -- improves test performance quite markedly. Furthermore, by employing an approximation suggested by \citet{brock2021high}, these gains come only at the cost of about $25\%$ extra compute.
\item The improvement brought by SAM often increases with less labeled training data, making SAM indispensable for data-limited tasks. We test this by subsampling the training splits of CBQA and SuperGLUE datasets at rates ranging from $2\%$ to $80\%$.
\end{enumerate}

\section{Related Works}
\paragraph{\textbf{Better Generalization.}} In light of flatter minima generalizing better, \citet{smith2017bayesian} showed that the inherent noise in SGD serves as a form of \emph{implicit regularization}, preventing the optimization from ever entering sharp valleys. Like SAM, entropy SGD~\citep{chaudhari2019entropy} \emph{explicitly} encourages flatter minima. \citet{smith2021origin,barrett2020implicit} analyzed SGD's generalization formally by way of continuous-time gradient flow.
Optimization routines based on adversarial risk~\citep{zhu2019freelb,he2020deberta} and trust regions~\citep{jiang2019smart,aghajanyan2020better} have been proposed and shown to improve generalization across settings.

While the number of methods which provide implicit or explicit regularization is overwhelmingly large, methods like early stopping, weight decay (or $\ell_2$-regularization), dropout~\citep{srivastava2014dropout}, teacher-student or self-distillation~\citep{hinton2015distilling,mobahi2020self}, label smoothing~\citep{muller2019does}, batch normalization~\citep{ioffe2015batch}, mixup~\citep{zhang2017mixup}, and data-augmentation more broadly are among the most widely used in practice. Marginalization of Bayesian neural networks, though challenging, has been shown to result in superior generalization in some settings~\citep{wilson2020bayesian,mackay1995probable}.

While first-order optimization via SGD has been the prevailing way of training neural networks due to its efficiency and effectiveness, alternative second-order methods like K-FAC~\citep{martens2015optimizing} and Shampoo~\citep{gupta2018shampoo} have slowly gained traction, often enabled by clever engineering to make them feasible at scale. Notably, \citet{anil2020scalable} presents a scalable implementation of Shampoo that provides significant convergence and wall-clock time improvements compared to first-order methods. They demonstrate superior performance on machine translation and language modeling.

\paragraph{\textbf{SAM.}}
While this is, to the best of our knowledge, the first work detailing the benefits of SAM for language tasks, there have been successful applications of SAM in the vision domain. Notably, \citet{chen2021vision} showed that convolution-free vision models like vision transformers (ViTs)~\citep{dosovitskiy2020image} and MLP-Mixers~\citep{tolstikhin2021mlp} suffer from sharp minima and that SAM indeed smooths their loss landscapes. They crucially show that ViTs and MLP-Mixers outperfom ResNets of similar and greater size on ImageNet \emph{without} the use of pre-training or data augmentations that would otherwise be necessary to achieve reasonable performance. They show that SAM induces sparsity in both architectures and leads to more perceptive attention maps in ViTs. They observe empirically that data augmentation and SAM are alike in that they both smooth the landscape on average, but the latter does so by explicitly controlling the worst-case curvature, whereas the former smooths over the directions induced by the augmentations. Furthermore, they observe that SAM encourages linearity with respect to the input, exhibiting an effect similar to that of mixup~\citep{zhang2017mixup}. Lastly, they show that SAM helps contrastive learning and that it enables better robustness on corrupted examples from ImageNet-C~\citep{hendrycks2019benchmarking} and ImageNet-R~\citep{hendrycks2021many}.

In a similar spirit, \citet{brock2021high} proposed speeding up SAM significantly by using fewer examples when computing the ascent step, a strategy which we employ in this work, and they were able to apply it to ResNet model variants to advance the state of the art on ImageNet without extra data.

Meanwhile, in an attempt to make SAM's radius $\rho$ invariant to the scale of the model parameters, \citet{kwon2021asam} proposed an adaptive version named Adaptive Sharpness-Aware Minimization (ASAM), which they then show empirically to outperform normal SAM on a set of benchmark vision tasks.

\section{Review of Sharpness-Aware Minimization~(SAM)}
We begin by briefly reviewing the SAM algorithm; interested readers can see the original paper for a thorough treatment. In our presentation, we use the $\ell_2$ norm ($p=2$ using notation from the original paper), assume a general optimizer (instead of vanilla SGD), and use the approximation proposed by \citet{brock2021high} to compute the ascent gradient (adversarial point) efficiently.
Given a loss function $L: \mathcal{W} \times \mathcal{X} \times \mathcal{Y} \rightarrow \mathbb{R}_+$, 
SAM seeks to find the parameter $w$ whose neighborhood has low training loss by optimizing the minimax objective:
\begin{align*}
    \min\limits_{w} \max\limits_{||\epsilon||_2 \leq \rho} L_\text{train}(w + \epsilon).
\end{align*}
Finding the exact optima $\epsilon^*$ of the inner-maximization is challenging, so \citet{foret2020sharpness} employ a first-order approximation, resulting in:
\begin{align*}
    \hat{\epsilon}(w) &= \operatorname*{argmin}\limits_{||\epsilon||_2 \leq \rho} L_\text{train}(w) + \epsilon^T \nabla_w L_\text{train}(w) \\
    & = \rho \nabla_w L_\text{train}(w)/||\nabla_w L_\text{train}(w)||_2.
\end{align*}
That is, $\hat{\epsilon}$ is just a scaling of the loss gradient at the current parameters. After computing $\hat{\epsilon}(w)$, SAM performs gradient descent using the gradient $\nabla_w L_\text{train}(w)|_{w_\text{adv}}$ at the nearby ``adversarial'' point $w_\text{adv}(w) \triangleq w + \hat{\epsilon}(w)$.

Put another way, SAM plugs-and-plays with any first-order optimizer by simply replacing the gradient of the mini-batch $\mathcal{B}$ at the current model weights $w_t\in \mathcal{W}$ with the gradient computed at $w_\text{adv}$. $w_\text{adv}$ itself is computed by taking a gradient \emph{ascent} step of size $\rho$ along the unit gradient vector $\nabla_w L_\mathcal{M}(w)/||\nabla_w L_\mathcal{M}(w)||_2|_{w_t}$, where $\mathcal{M}$ can be the mini-batch $\mathcal{B}$, or a subset of it for enhanced efficiency. We found that setting $\mathcal{M}$ to be $1/4$-th of $\mathcal{B}$ sped up the method significantly with little loss in quality, in line with the recommendation of \citet{brock2021high}. The end-to-end algorithm is outlined in Algorithm~\ref{alg:main}.

        \begin{algorithm}[!t]
        \caption{\label{alg:main} Efficient SAM Algorithm.}
        \begin{algorithmic}[1]
        \STATE \textbf{input:} training set $\mathcal{S} \triangleq \cup_{i=1}^n\{(x_i, y_i)\}$, loss function $L:\mathcal{W} \times \mathcal{X} \times \mathcal{Y} \rightarrow \mathbb{R}_+$, batch size $b$, neighborhood size $\rho>0$ (default $0.15)$, ascent micro-batch size $a \leq b$ (default $b/4$), first-order optimizer update $\operatorname{opt}: \mathcal{W} \times \mathcal{W} \rightarrow \mathcal{W}$.
        \STATE initialize parameters $w_0$, $t=0$.
         \WHILE{not converged}
          \STATE sample batch $\mathcal{B} = \{(x_{1}, y_{1}), ... ,(x_{b}, y_{b})\}$.
          \STATE sample ascent micro-batch $\mathcal{M} = \{(x_{1}, y_{1}), ... , (x_{a}, y_{a})\}$.
          \STATE compute adversarial (ascent) point: $w_\text{adv} = w_t + \rho\frac{\nabla_w L_\mathcal{M}(w)}{||\nabla_w L_\mathcal{M}(w)||_2}|_{w_t}$.
          \STATE compute gradient approximation for the SAM objective: $g_\text{adv} = \nabla_w L_\mathcal{B}(w)|_{w_\text{adv}}$.
          \STATE update parameters: $w_{t+1} = \operatorname{opt}(w_t, g_\text{adv})$.
          \STATE $t=t+1$.
          \ENDWHILE
         \RETURN{$w_t$}
         \label{alg:sam-algorithm}
        \end{algorithmic}
        \end{algorithm}

\section{Experiments}
With SAM reviewed, we now discuss our experiments. We evaluate SAM on a range of natural language understanding tasks using the T5 (text-to-text Transformer) framework~\citep{raffel2019exploring}. T5 casts NLU tasks as sequence-to-sequence ones that are learned using an encoder-decoder Transformer~\citep{vaswani2017attention} architecture setup. These Transformer models are typically pre-trained on large corpora, like the Colossal Clean Crawled Corpus (C4)~\citep{raffel2019exploring}, with, for example, the objective of predicting a short contiguous span of text that was intentionally corrupted in a snippet of input text. The pre-trained model is typically fine-tuned on a single task or a mixture of multiple tasks, the latter enabled by the fact that the framework treats all tasks as simple input-to-target sequence predictions.

To this end, we evaluate SAM in two ways:
\begin{enumerate}
\item When publicly available pre-trained checkpoints of the T5.1.1 model variant are fine-tuned with and without SAM, on SuperGLUE, GLUE, TyDiQA, and the Closed-Book Question Answering benchmarks: Web Questions, Natural Questions, TriviaQA. We show SAM improves generalization across benchmarks and four model sizes: Small (77M parameters), Base (250M), Large (800M), and XL (3B).
\item To show how it helps when task data is limited, we report results when the training splits of these benchmarks at various rates, ranging from $2\%$ to $80\%$.

\end{enumerate}
\begin{table*}[t]
\small
    \centering
    \begin{tabular}{l|c|ccccccccc}
    \toprule
        Model &	SGlue &	BoolQ &	CB & 	CoPA & MultiRC &	ReCoRD &	RTE & 	WiC &	WSC \\ 
        \midrule
Small & 67.7 & 72.6 & 89.4 / \bf89.3 & \bf67.0 & \textbf{68.5} / 21.4 & 61.7 / 60.8 & 69.3 & 65.4 & 72.1 \\
Small + SAM (0.05) & \bf68.4 & \bf73.5 & \bf92.1 / \bf89.3 & 61.0 & \bf68.5 / \bf22.8 & \bf62.1 / \bf61.0 & \bf69.7 & \bf65.7 & \bf79.8 \\
\midrule
Base & 75.3 & 80.0 & 91.7 / \bf94.6 & 71.0 & 75.4 / 35.4 & 76.2 / 75.4 & 80.9 & 69.3 & 76.9 \\
Base + SAM (0.15) & \bf78.5 & \bf82.2 & \bf93.7 / \bf94.6 & \bf78.0 & \bf77.5 / \bf39.1 & \bf78.2 / \bf77.2 & \bf85.9 & \bf70.4 & \bf81.7 \\
\midrule
Large & 84.3 & 86.6 & \bf99.4 / \bf98.2 & \bf89.0 & 83.7 / 51.0 & 86.5 / 85.6 & \bf89.2 & 72.9 & 84.6 \\
Large + SAM (0.15) & \bf84.6 & \bf88.0 & 95.0 / 96.4 & 86.0 & \bf84.0 / \bf53.7 & \bf87.3 / \bf86.4 & \bf89.2 & \bf75.2 & \bf86.5 \\
\midrule
XL & 87.2 & 88.6 & 93.7 / 96.4 & \bf95.0 & 86.9 / 61.1 & 89.5 / 88.4 & 91.3 & 74.9 & 89.4 \\
XL + SAM (0.15) & \bf89.1 & \bf89.4 & \bf100.0 / \bf100.0 & \bf95.0 & \bf87.9 / \bf63.7 & \bf90.9 / \bf90.0 & \bf92.1 & \bf75.5 & \bf94.2 \\
        \bottomrule
    \end{tabular}
    \caption{Experimental results (dev scores) on the (full) SuperGLUE benchmark. Public checkpoints of various sizes are fine-tuned with and without SAM for 250k steps. We see that SAM improves performance across \emph{all} model sizes.}
    \label{tab:finetune_superglue}
\end{table*}

\begin{table*}[t]
    \small
    \centering
    \begin{tabular}{l|c|ccccccccc}
    \toprule
        Model &	Glue &	CoLA& SST& MRPC	& STSB & QQP & MNLI & QNLI& RTE \\ 
        \midrule
Small & 79.8 & 39.9 & \bf92.3 & 90.1 / 85.8 & 85.4 / 84.9 & 87.3 / 90.5 & 80.9 / \bf81.4 & 87.6 & \bf74.0 \\
Small + SAM (0.05) & \bf79.9 & \bf42.9 & 92.1 & \bf90.9 / \bf87.3 & \bf85.5 / \bf85.3 & \bf87.6 / \bf90.7 & \bf81.0 / 81.4 & \bf87.7 & 70.4 \\
\midrule
Base & 84.7 & \bf50.9 & \bf94.2 & 91.5 / 88.2 & 88.5 / 88.3 & 88.5 / 91.4 & \bf87.3 / 87.6 & 92.1 & 81.6 \\
Base + SAM (0.15) & \bf85.1 & 49.8 & \bf94.2 & \bf93.4 / \bf90.7 & \bf90.0 / 89.7 & \bf89.0 / 91.7 & \textbf{87.3} / 87.5 & \bf92.5 & \bf82.7 \\
\midrule
Large & 86.2 & 57.8 & 95.3 & 89.5 / 85.3 & 88.3 / 88.5 & 87.8 / 90.9 & 88.9 / 89.0 & 93.4 & 85.9 \\
Large + SAM (0.15) & \bf88.3 & \bf65.6 & \bf95.6 & \bf93.6 / 91.2 & \bf90.2 / 89.8 & \bf89.7 / 92.2 & \bf89.9 / 89.7 & \bf94.5 & \bf85.9 \\
\midrule
XL & 89.9 & \bf70.7 & 96.3 & 92.8 / 90.0 & 91.2 / 91.0 & 89.6 / 92.1 & 91.1 / 91.3 & 95.5 & 90.6 \\
XL + SAM (0.15) & \bf90.2 & 69.7 & \bf96.9 & \bf93.2 / 90.7 & \bf91.7 / 91.8 & \bf90.2 / 92.7 & \bf91.4 / 91.5 & \bf95.8 & \bf91.0 \\
\bottomrule
    \end{tabular}
    \caption{Experimental results (dev scores) on the (full) GLUE benchmark. Public checkpoints of various sizes are fine-tuned with and without SAM on the mixture of tasks for 250k steps. We see that SAM improves performance across \emph{all} model sizes.}
    \label{tab:finetune_glue}
\end{table*}
\begin{table}[!t]
    \centering
    \begin{tabular}{l|c}
    \toprule
        Model &	avg. (F1/EM) \\
\midrule
Small & 73.4 / 62.1 \\
Small + SAM (0.02) & \bf74.3 / \bf63.0 \\
\midrule
Base & 81.6 / 71.0 \\
Base + SAM (0.02) & \bf82.0 / \bf71.3 \\
\midrule
Large & 85.6 / 75.3 \\
Large + SAM (0.02) & \bf85.9 / \bf76.1 \\
\midrule
XL & 87.0 / 77.4 \\
XL + SAM (0.05) & \bf87.3 / \bf77.7 \\
        \bottomrule
    \end{tabular}
    \caption{Average results on TyDiQA-GoldP. Public checkpoints of mT5 of various sizes are fine-tuned on TyDiQA-GoldP for 20k steps. SAM boosts performance acros model sizes here as well. We found that smaller values of $\rho$ than those used for the English-only T5 model were necessary to achieve good performance here. Full, per-language results are shown in the Appendix.}
    \label{tab:tydiqa_overall}
\end{table}

\begin{table*}[t]
    \centering
    \begin{tabular}{l|ccc}
    \toprule
    Model &	Natural Q. & Web Q. & TriviaQA \\
        \midrule
Small & 16.7 / 12.4 & 22.8 / 16.5 & 10.2 / 7.3 \\
Small + SAM (0.05) & \bf17.5 / 13.1 & \bf23.5 / 16.9 & \bf11.0 / 7.8 \\
\midrule
Base & 23.2 / 18.1 & 29.7 / 22.5 & 19.3 / 15.3 \\
Base + SAM (0.15) & \bf25.7 / 20.6 & \bf31.0 / 24.5 & \bf21.5 / 17.4 \\
\midrule
Large & 27.4 / 22.3 & 34.3 / 27.6 & 25.2 / 20.9 \\
Large + SAM (0.15) & \bf30.6 / 25.0 & \bf36.4 / 29.6 & \bf28.5 / 24.2 \\
\midrule
XL & 33.5 / 27.5 & 39.3 / 31.6 & 36.5 / 31.1 \\
XL + SAM (0.15) & \bf34.7 / 28.8 & \bf40.7 / 33.3 & \bf38.0 / 32.6 \\
\bottomrule
    \end{tabular}
    \caption{Experimental results (F1/EM) (test scores) on the (full) CBQA tasks. Public checkpoints of various sizes are fine-tuned with and without SAM on the mixture of tasks for 20k steps. We see that SAM improves performance across \emph{all} model sizes.}
    \label{tab:finetune_cbqa}
\end{table*}

\begin{table*}[t]
    \centering
    \begin{tabular}{l|ccc}
    \toprule
    Model &	Natural Q. & Web Q. & TriviaQA \\
        \midrule
Small & 19.2 / 15.0 & 23.8 / 17.7 & 10.9 / 8.1 \\
Small + SAM (0.05) & \bf20.8 / 16.5 & \bf25.9 / 20.4 & \bf12.3 / 9.4 \\
\midrule
Base & 26.3 / 21.1 & 31.6 / 26.0 & 20.6 / 16.8 \\
Base + SAM (0.15) & \bf27.8 / 22.6 & \bf33.6 / 27.5 & \bf23.7 / 19.5 \\
\midrule
Large & 28.1 / 23.0 & 32.7 / 25.8 & 25.1 / 20.8 \\
Large + SAM (0.15) & \bf30.8 / 25.3 & \bf34.4 / 28.0 & \bf28.8 / 24.2 \\
\midrule
XL & 33.4 / 27.3 & 37.1 / 30.6 & 35.5 / 30.2 \\
XL + SAM (0.15) & \bf34.2 / 28.3 & \bf39.4 / 32.3 & \bf37.6 / 32.0 \\
\bottomrule
    \end{tabular}
    \caption{Experimental results (F1/EM) (test scores) on the (full) CBQA tasks, where the model is trained on each of the three tasks separately, rather than on a mixture. We see that SAM improves performance across \emph{all} model sizes, as we observed when training on the mixtures. This suggests that SAM's gains are not solely due to some ability to better leverage multi-task learning.}
    \label{tab:finetune_cbqa_ablation}
\end{table*}

\begin{figure*}[t]
\centering
\includegraphics[width=0.32\textwidth]{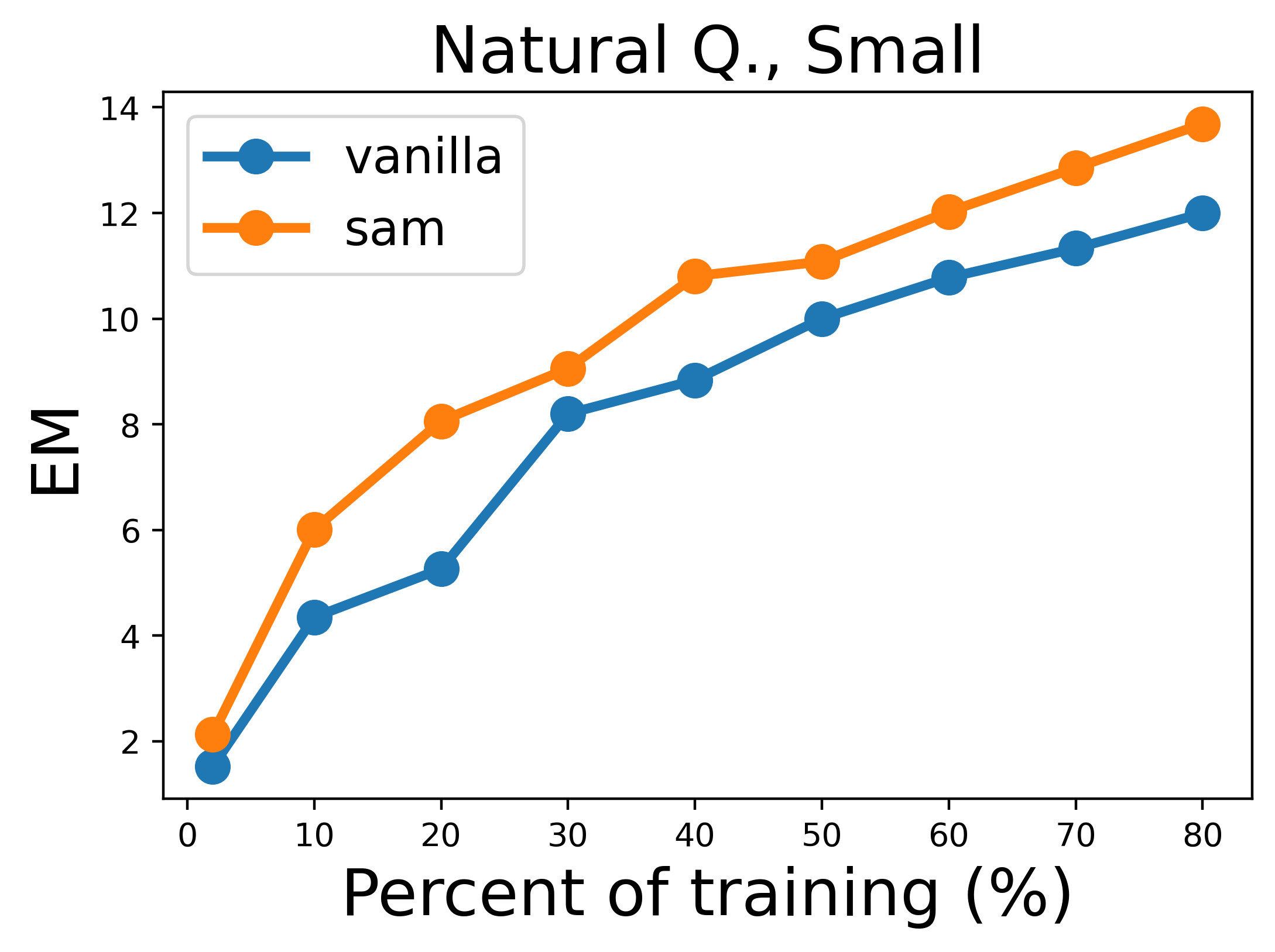}
\includegraphics[width=0.32\textwidth]{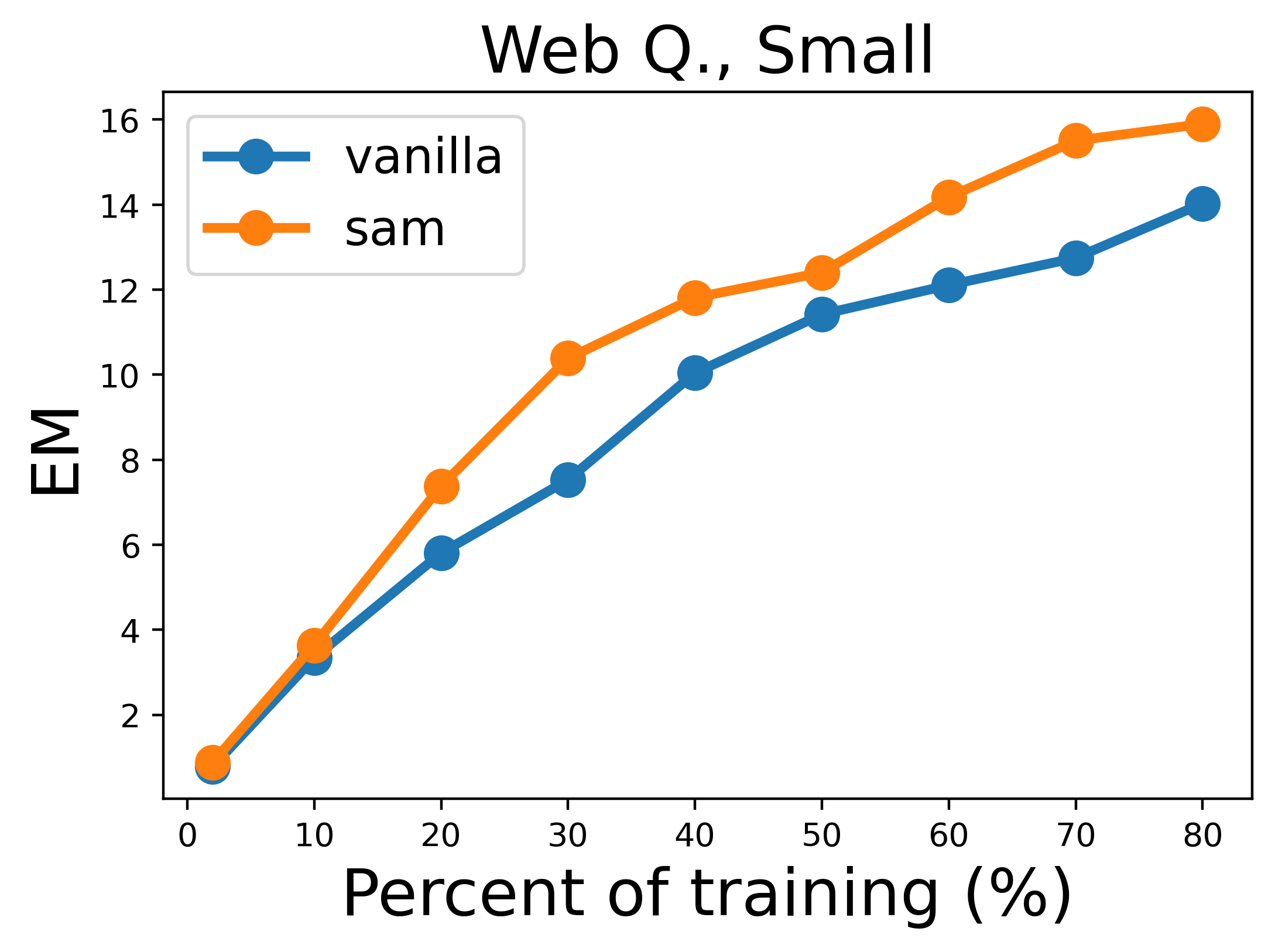}
\includegraphics[width=0.32\textwidth]{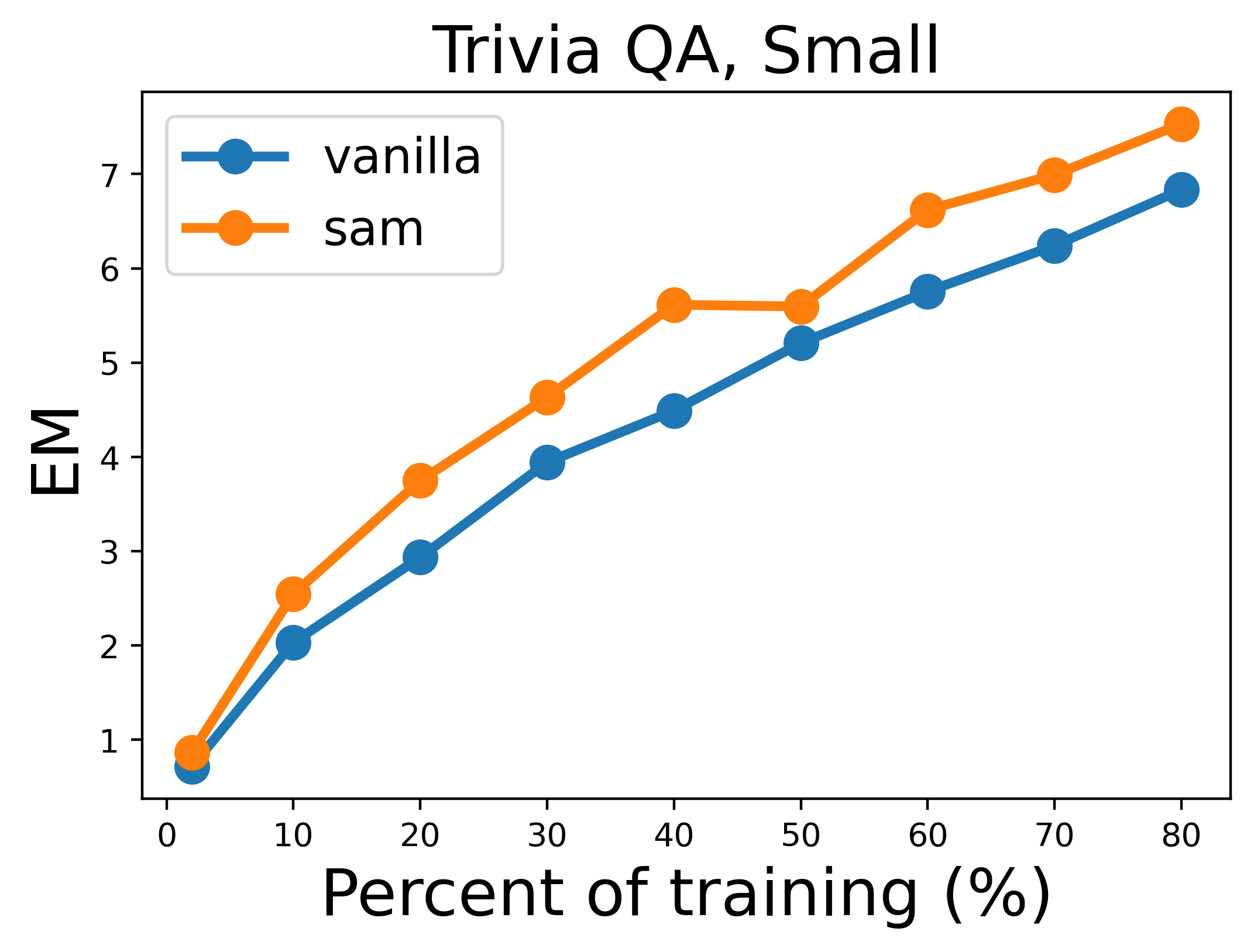} \\
\includegraphics[width=0.32\textwidth]{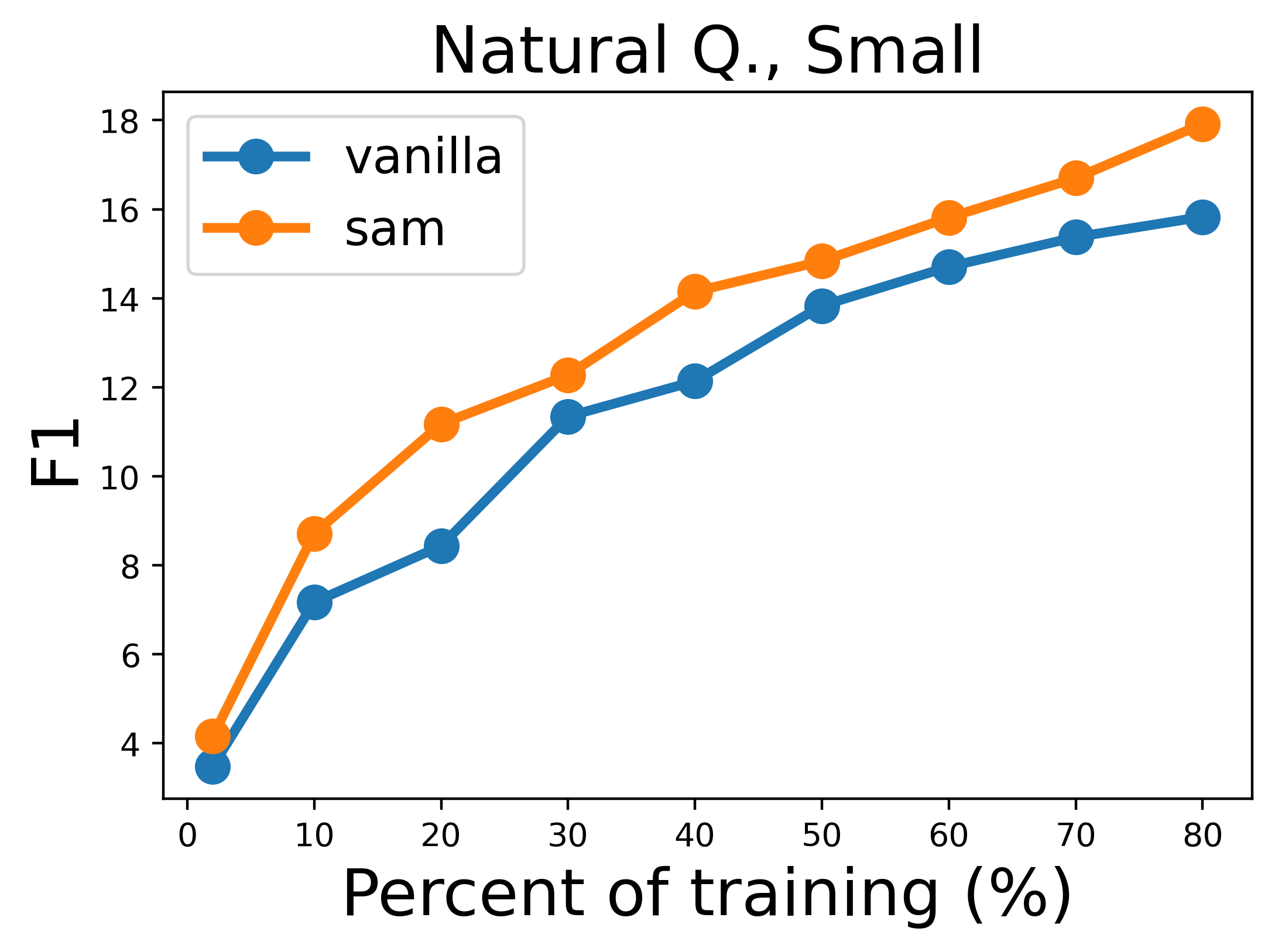}
\includegraphics[width=0.32\textwidth]{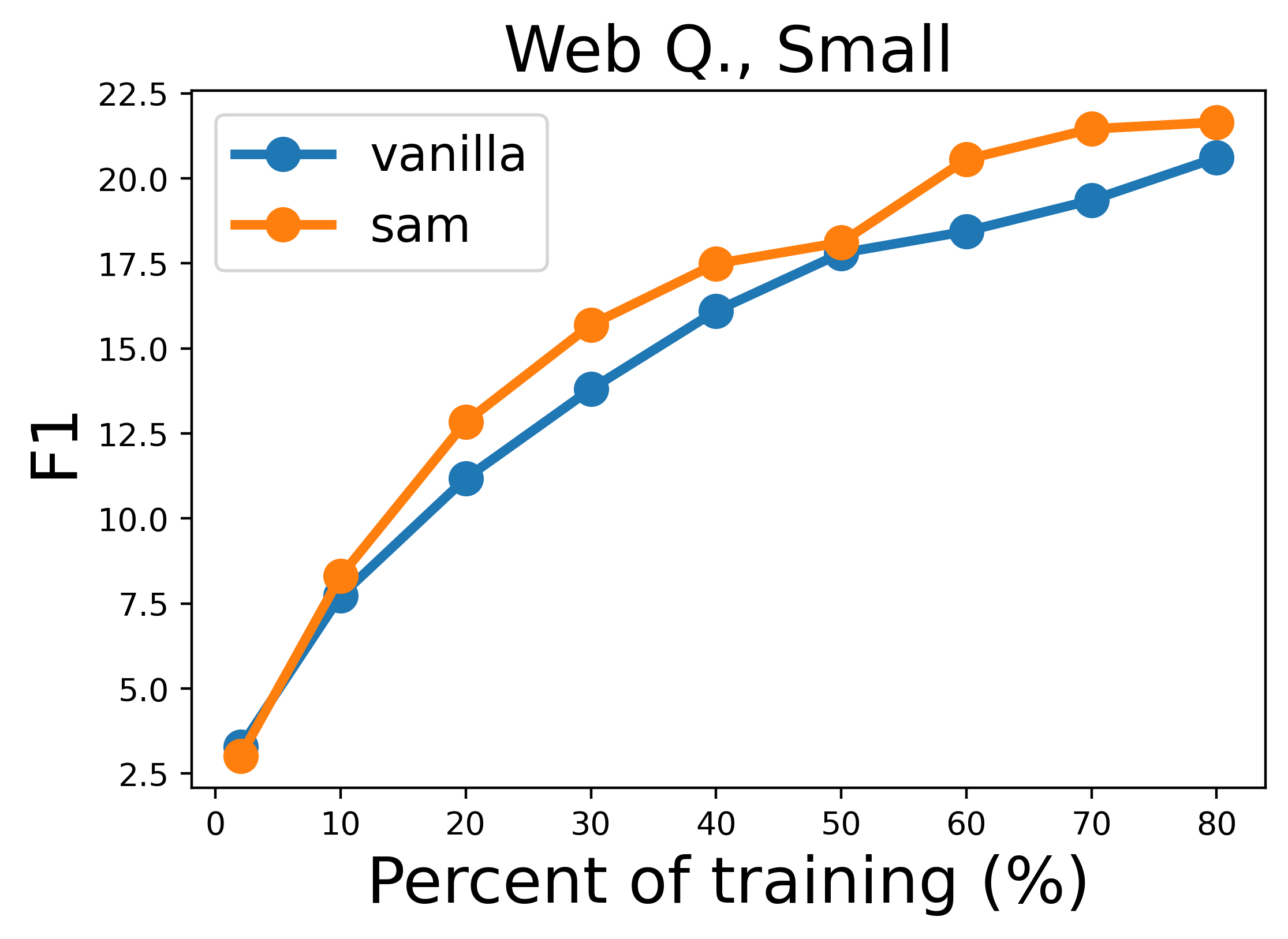}
\includegraphics[width=0.32\textwidth]{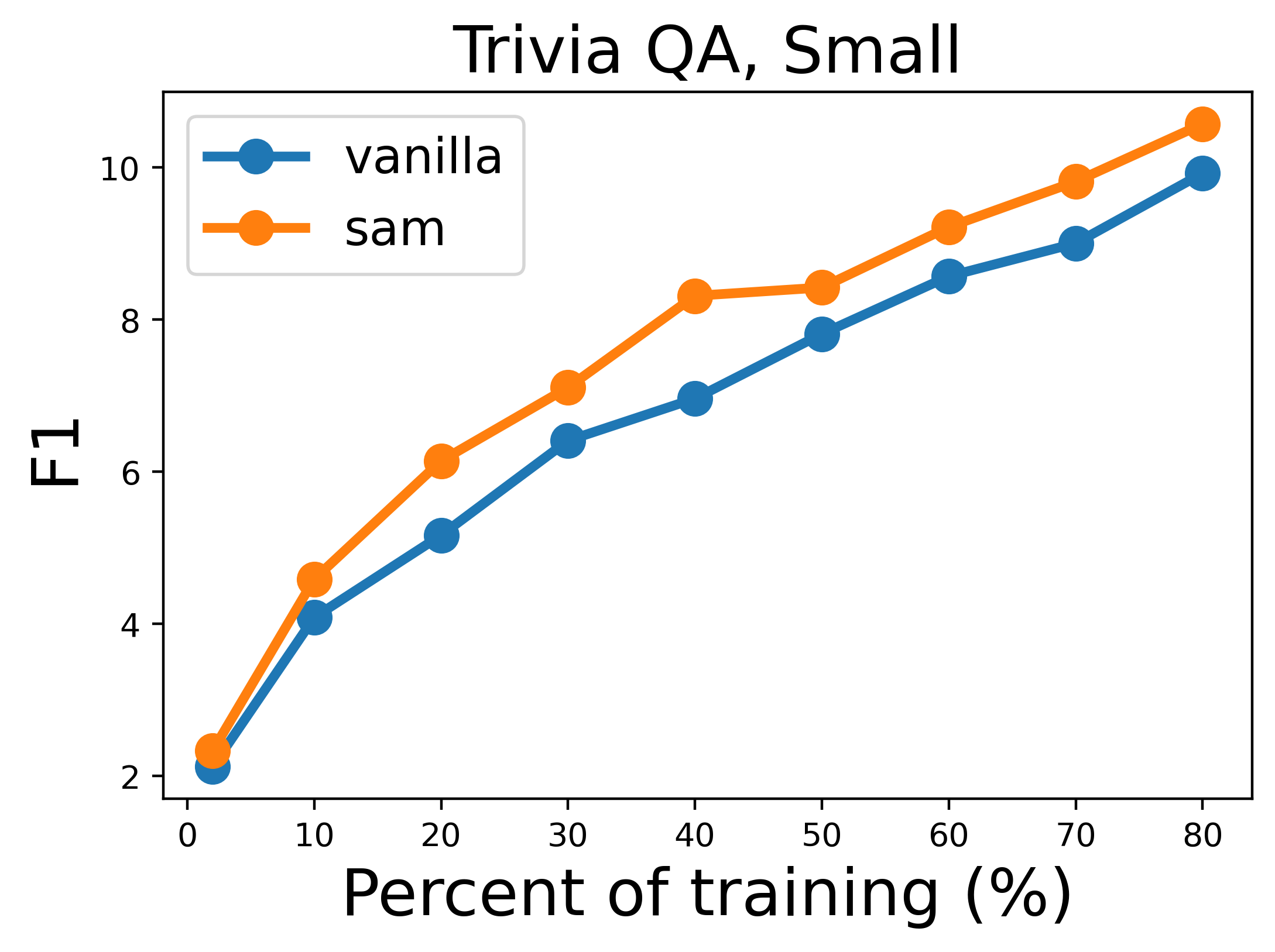} \\
\includegraphics[width=0.32\textwidth]{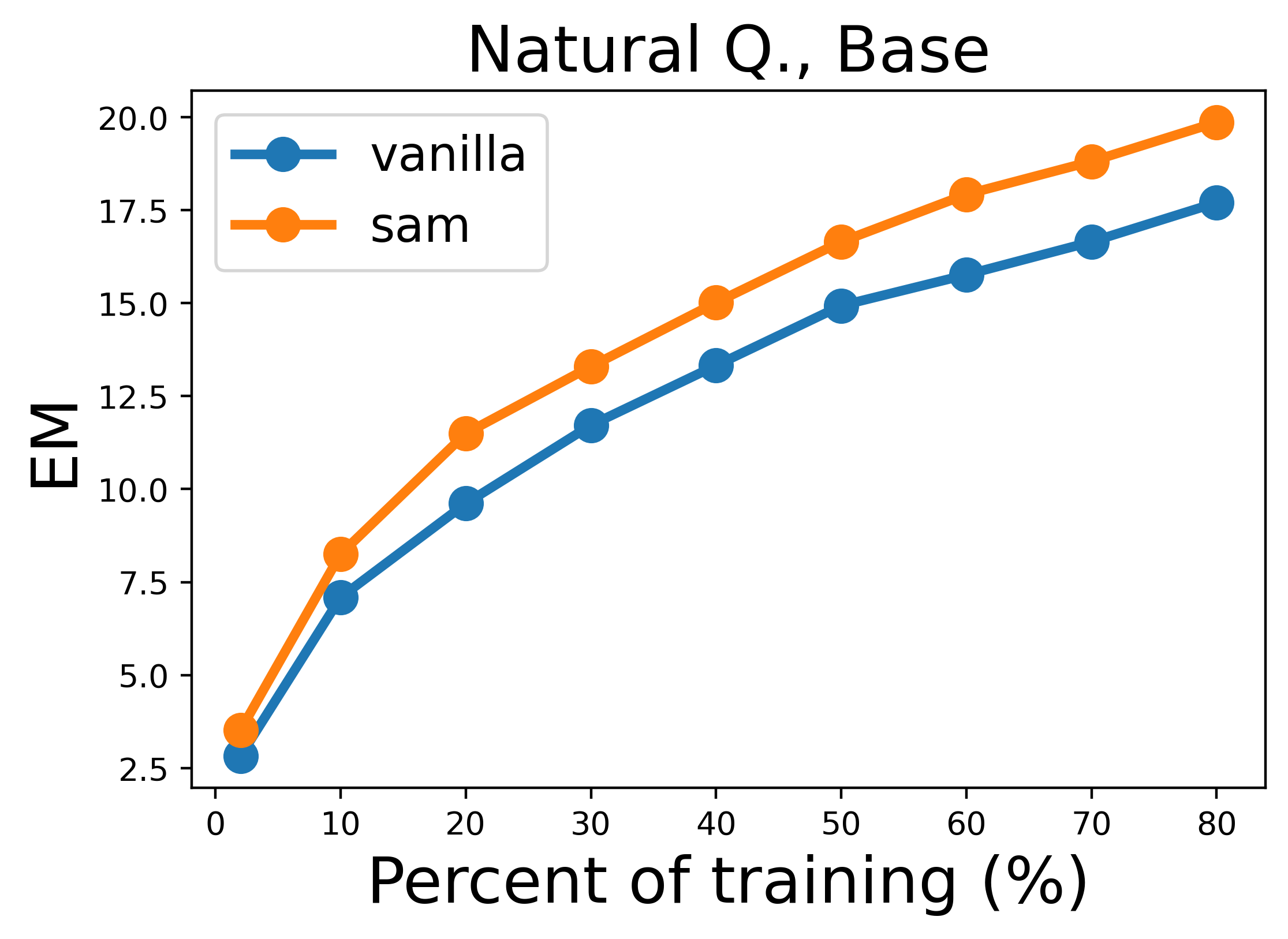}
\includegraphics[width=0.32\textwidth]{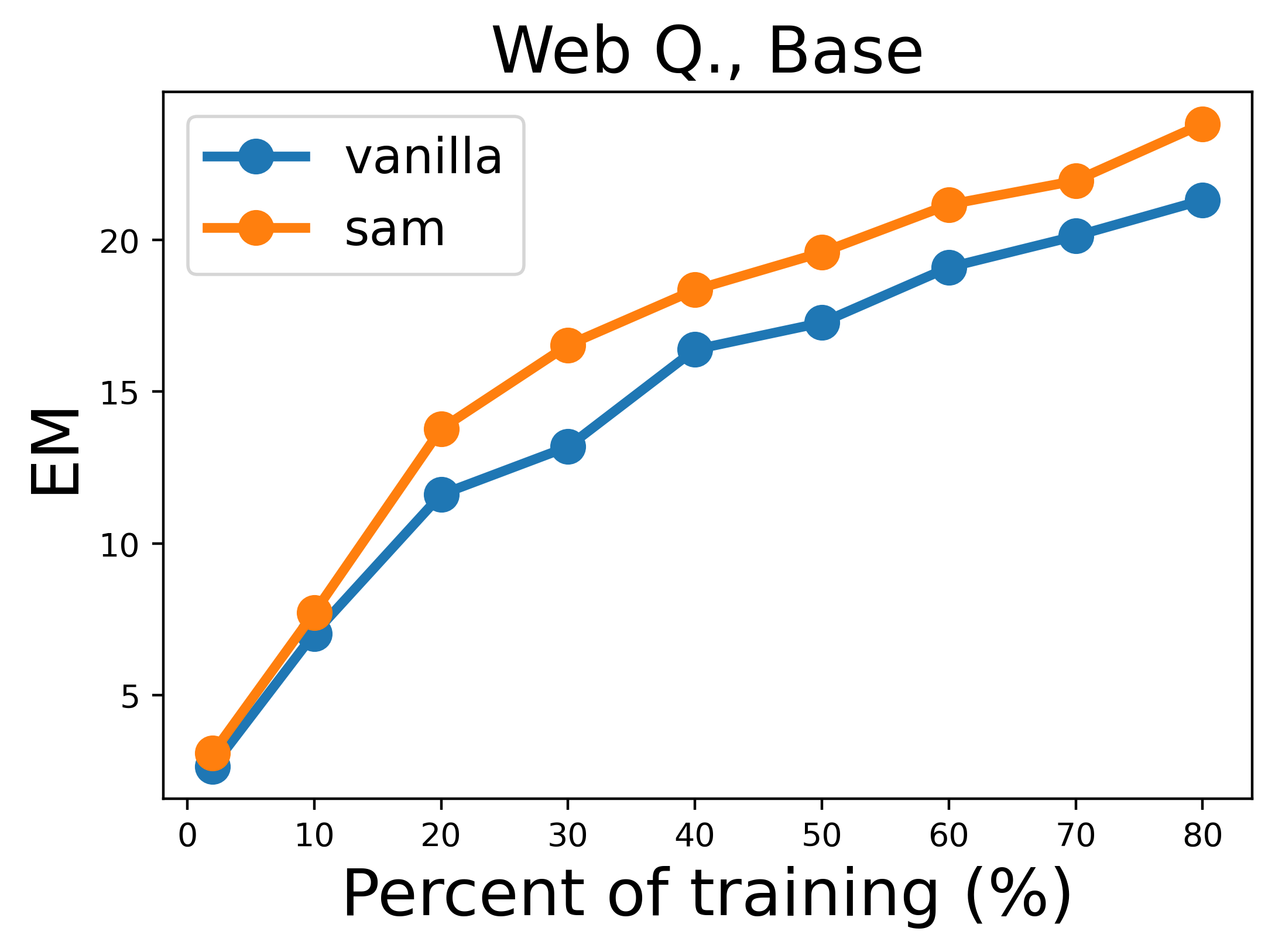}
\includegraphics[width=0.32\textwidth]{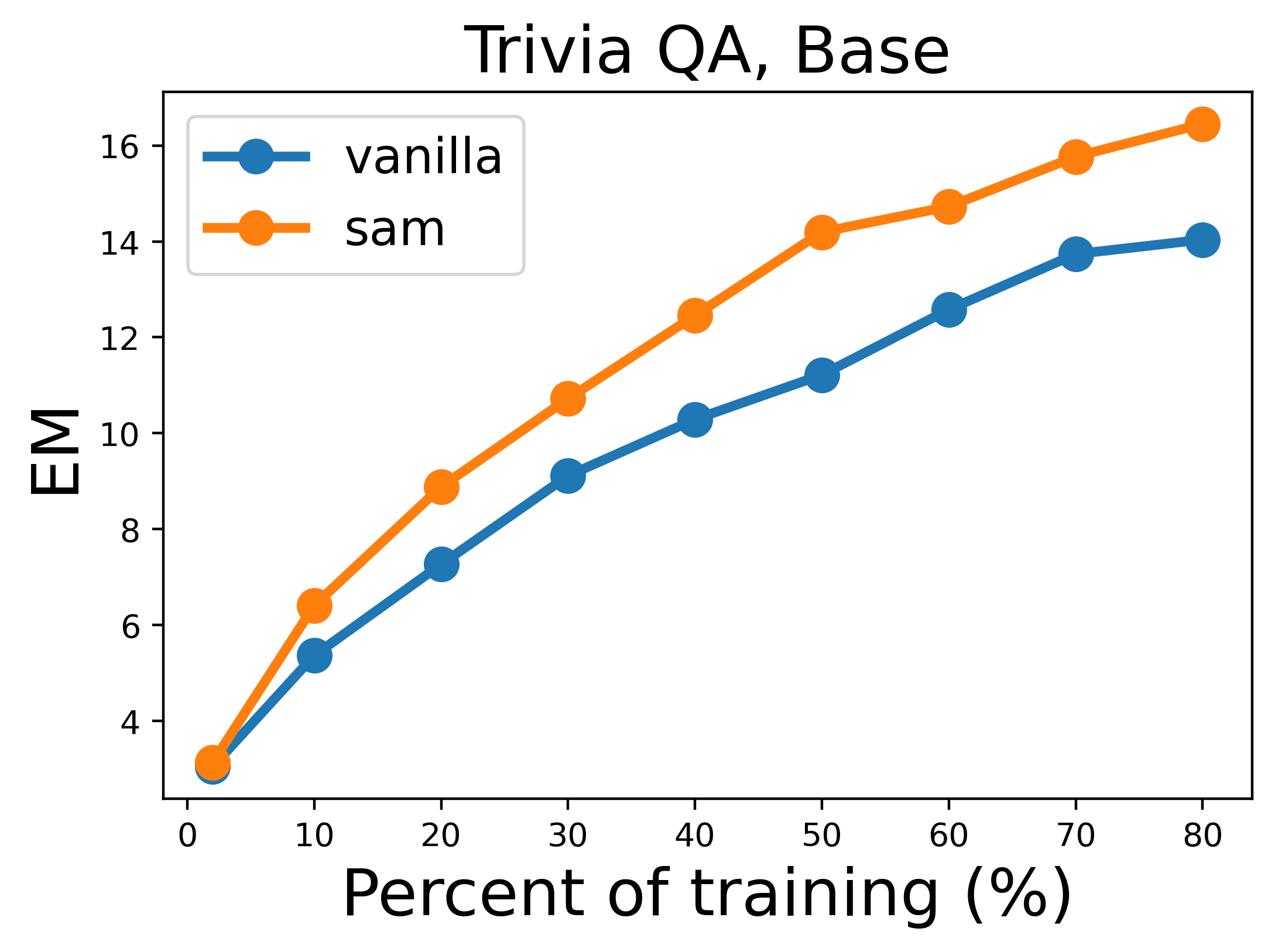} \\
\includegraphics[width=0.32\textwidth]{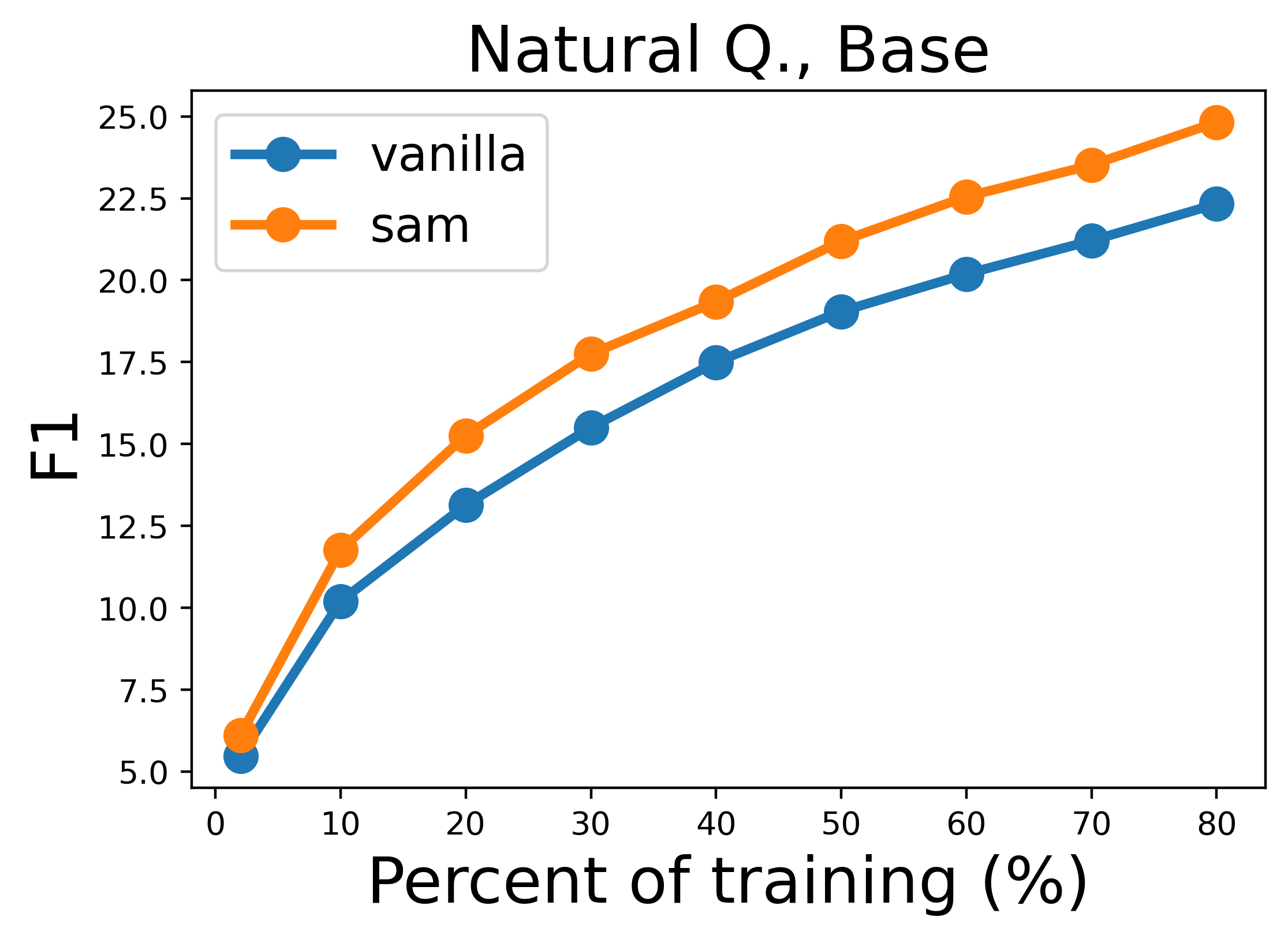}
\includegraphics[width=0.32\textwidth]{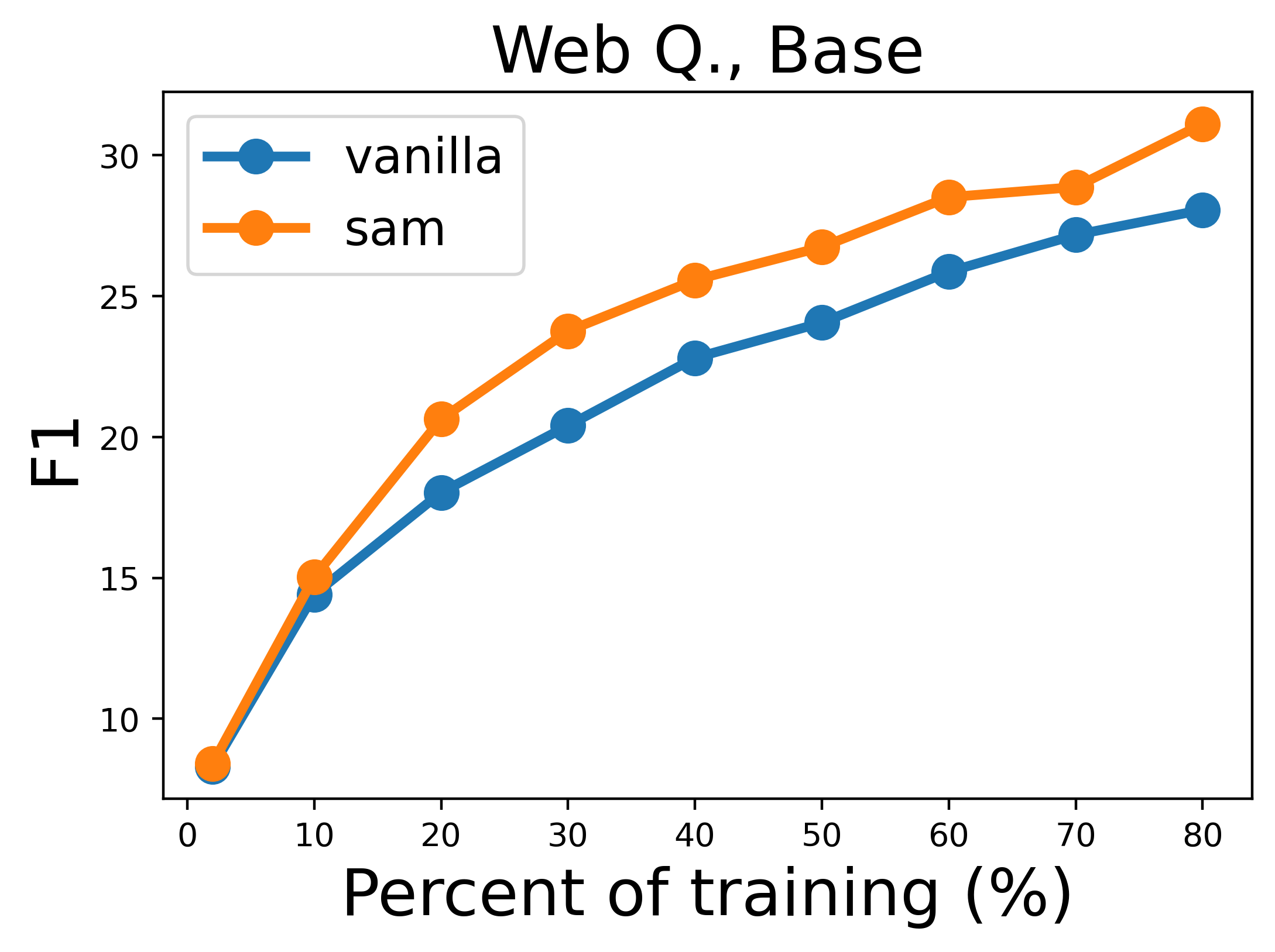}
\includegraphics[width=0.32\textwidth]{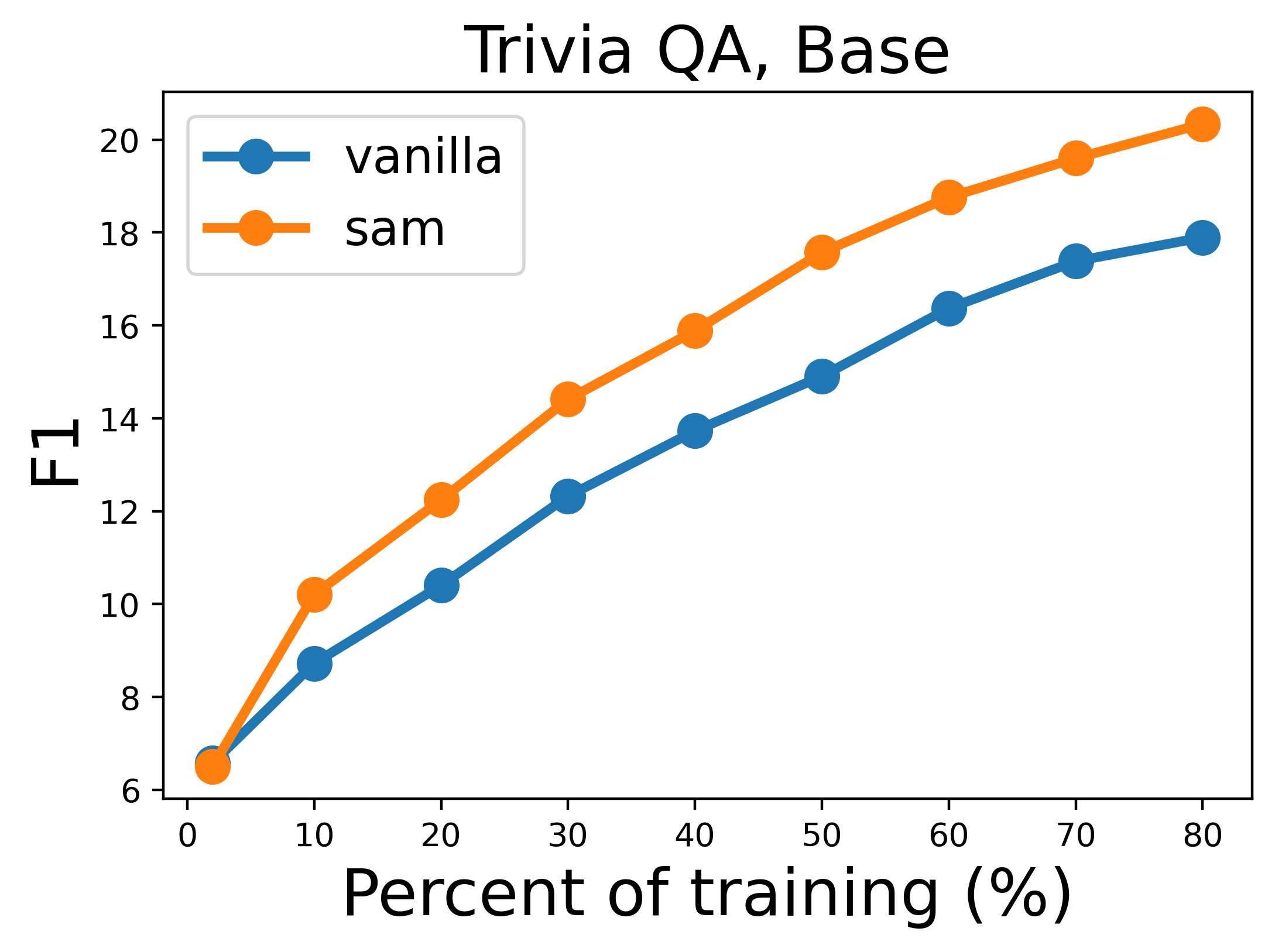} \\
\caption{CBQA results at various training data sampling rates, for the Small (\textbf{top half}) and Base (\textbf{bottom half}) models. We see that SAM's improvement is consistent across data size regimes and that the relative improvement is often largest in the ballpark of 20\%.}
\label{fig:data-limited}
\end{figure*}

\begin{table*}[t]
\small
    \centering
    \begin{tabular}{l|c|ccccccccc}
    \toprule
        Model &	SGlue &	BoolQ &	CB & 	CoPA & MultiRC &	ReCoRD &	RTE & 	WiC &	WSC \\ 
        \midrule
Small & 50.2 & \bf60.8 & 37.0 / 55.4 & 52.0 & 60.1 / 11.0 & 33.9 / 32.5 & \bf54.5 & 54.4 & 65.4 \\
Small + SAM (0.05) & \bf51.9 & 60.5 & \bf45.6 / 66.1 & \bf53.0 & \bf61.1 / 12.5 & \bf36.7 / 34.5 & 52.3 & \bf55.2 & \bf66.3 \\
\midrule
Base & 52.9 & 59.6 & 32.3 / 55.4 & 53.0 & 60.2 / 11.8 & 47.8 / 46.5 & 58.5 & \bf57.2 & \bf68.3 \\
Base + SAM (0.15) & \bf56.7 & \bf61.4 & \bf41.8 / 64.3 & \bf55.0 & \bf62.4 / 15.7 & \bf59.7 / 57.9 & \bf62.5 & 55.3 & \bf68.3 \\
\midrule
Large & 62.8 & 65.3 & 40.1 / 62.5 & \bf62.0 & \bf71.6 / 24.0 & 80.4 / 78.9 & \bf69.7 & 57.7 & \bf69.2 \\
Large + SAM (0.15) & \bf64.3 & \bf77.3 & \bf47.9 / 69.6 & 59.0 & 69.0 / 20.4 & \bf81.5 / 80.0 & 65.0 & \bf59.6 & \bf69.2 \\
\midrule
XL & 75.9 & \bf84.5 & 57.0 / 82.1 & \bf86.0 & \bf82.4 / 48.7 & 83.3 / 81.7 & 78.7 & \bf66.0 & 74.0 \\
XL + SAM (0.15) & \bf77.0 & 82.5 & \bf58.9 / 83.9 & 85.0 & 79.9 / 45.3 & \bf86.8 / 85.6 & \bf80.5 & 64.4 & \bf83.7 \\
        \bottomrule
    \end{tabular}
    \caption{SuperGLUE results when only 5\% of the training data is available. We see again that SAM boosts performance across the board, adding a whopping $7.2\%$ relative improvement on the Base model.}
    \label{tab:finetune_superglue_5}
\end{table*}

\begin{table*}[t]
    \centering
    \begin{tabular}{l|ccc}
    \toprule
    Model &	Natural Q. & Web Q. & TriviaQA \\
        \midrule
Small & 4.9 / 2.9 & 4.8 / 1.8 & 3.0 / 1.3 \\
Small + SAM (0.05) & \bf6.0 / 3.7 & \bf7.1 / 2.2 & \bf3.3 / 1.6 \\
\midrule
Base & 7.9 / 4.8 & \textbf{13.7} / 4.6 & 7.6 / 4.2 \\
Base + SAM (0.15) & \bf8.6 / 5.6 & 12.2 / \textbf{5.7} & \bf7.7 / 4.4 \\
\midrule
Large & 8.7 / 5.2 & 14.0 / 7.0 & 9.8 / 6.0 \\
Large + SAM (0.15) & \bf10.5 / 6.6 & \bf14.9 / 7.7 & \bf10.6 / 7.1 \\
\midrule
XL & 13.1 / 8.0 & 20.6 / 11.9 & \bf{19.6} / 15.3 \\
XL + SAM (0.15) & \bf13.4 / 8.1 & \bf22.9 / 13.6 & 19.1 / 14.5 \\
\bottomrule
    \end{tabular}
    \caption{CBQA results when only 5\% of the training data is available. We see SAM helps here as it did for subsampled SuperGLUE. For Natural Questions, SAM improves Base model performance by a relative $8.86\% / 16.6\%$ (F1/EM).}
    \label{tab:finetune_cbqa_5}
\end{table*}

\subsection{Setup}
\paragraph{Framework.}
For all experiments, we train using Jax~\citep{jax2018github} and Google Cloud TPUs. To ensure fair comparisons, eliminate the impact of exogenous factors, and reduce the possibility of software bugs, we train both standard and SAM-enabled models using the same codebase and settings, so that the code paths are identical except for the gradient calculation at each step, wherein SAM behaves differently. Our implementation of SAM is an adaptation of an existing open-source implementation\footnote{\url{https://github.com/google-research/sam}} to fit our framework for training language models.
\paragraph{Efficient SAM.}
In \citet{foret2020sharpness}, the idea of partitioning the ascent mini-batch into $m$ disjoint micro-batches and computing a distinct adversarial point for each micro-batch and then averaging the SAM-gradients at each of these points was proposed under the name $m$-sharpness. It was noted there and in follow-up work~\citep{chen2021vision} that $m>1$ can result in better performance. This modification incurs $m$-times more compute under a naive sequential implementation (though it can be parallelized well if multiple devices are available).

Meanwhile, \citet{brock2021high} suggests (in the Appendix) using roughly 20\% of the examples from the mini-batch for computing the adversarial point, observing little loss in model quality. With $m=1$, this approximation roughly reduces SAM's relative runtime from 2x to $1.2$x. Since we understand how a $2m$-x slow-down of model training may be prohibitive or significantly deter SAM's widespread adoption, we, at the possible loss of larger improvements, set $m = 1$ and use 1/4-th (25\%) of the mini-batch, or the number of available training devices (TPU cores in our case), whichever is larger, to compute SAM's adversarial point. This is necessary because the mini-batch gradient computation is parallelized over devices and each device must receive at least one example. We've observed from wall-clock times that with these settings, SAM is all in all about 25\% slower than standard training.

\paragraph{Hyper-parameters.}
SAM has a single hyper-parameter $\rho$, which is size of the step taken along the unit adversarial gradient vector. We search the range $[0.02, 0.05, 0.1, 0.15, 0.2, 0.3]$ a \emph{single} time \emph{only} when fine-tuning on SuperGLUE. We found that $0.05$ is a reasonable choice for T5.1.1 small models, and $0.15$ for the Base, Large, and XL variants, and so for all subsequent experiments except for TyDiQA, we use these choices without additional tuning. For the mT5 model on TyDiQA, we found that a smaller $\rho$ was necessary for good performance. For this, we searched the range $[0.01, 0.02, 0.05]$.

For all fine-tuning, we use the AdaFactor optimizer with learning rate 1e-3, 128 batch size, and the T5.1.1 settings.
For SuperGLUE, we use 10\% dropout rate, 512 input sequence length, 62 target sequence length, and fine-tune for 250k steps. For Natural Questions, Web Questions, and TriviaQA, we use 5\% dropout, 38 input sequence length, 18 target sequence length, and fine-tune for 20k steps. For TyDiQA, we use the official, public mT5 checkpoints, 10\% dropout, 1024 input sequence length, 512 target sequence length, and fine-tune for 20k steps. We run each experiment once, due to resource constraints, and we take the best checkpoint (stored every 1k steps for SuperGLUE and GLUE and every 200 steps for all other datasets) across training steps. Following standard practice, we report the best checkpoint for each task-metric pair (e.g. SuperGLUE CB F1) individually.

\subsection{Full Data Results}

Results for SuperGLUE and GLUE are shown in Table~\ref{tab:finetune_superglue} and Table~\ref{tab:finetune_glue} respectively. We observe that SAM improves the overall scores for both benchmarks across all T5 model sizes. For Base and XL sizes on SuperGLUE, SAM brings 4.2\% and 2.1\% relative gains in overall score respectively, while the gain for Large on GLUE is 2.4\%.
As shown in Table~\ref{tab:finetune_cbqa}, on Natural Questions, Web Questions, and Trivia QA tasks, we observe improvements for each task, metric (F1 and EM), and model size. For Base, we see a 13.8\%, 8.8\%, and 13.7\% gain on the exact match metric for Natural Questions, Web Questions, and Trivia QA respectively. For Large, these figures are 12.1\%, 7.2\%, and 15.7\%. Table~\ref{tab:tydiqa_overall} shows the results for TyDiQA-GoldP. Here, we observe more modest improvements in the 1-2\% range.
\paragraph{SAM improves performance on all model sizes.} In light of the conventional wisdom that ``larger models generalize better,'' we suspected, a priori, that SAM would be more helpful for the smaller models we consider, like Small and Base, and that we should expect substantial diminishing returns as we scale up the model size. Surprisingly, we did not observe any clear pattern with regards to size: indeed, sometimes the gains on XL were larger than those on Small. Thus, we lean to recommend SAM to all practitioners regardless of the regime in model capacity they are working in.
\paragraph{SAM improves single-task and multi-task learning alike.} Thus far, SAM has been trained on a mixture of tasks, where the influence of a particular task is proportional to the number of examples in its training split (i.e. no artificial up or down-weighting). To rule out the possibility that the gains observed are \emph{solely} due to some ability of SAM's to leverage multi-task learning and improve cross-task transfer, we conduct the following ablation. For each of the three CBQA tasks, we train only on a single task and report the performance on that task's test set. Results are shown in Table~\ref{tab:finetune_cbqa_ablation}. Indeed, we see similar gains when training and testing on each single task individually. We conclude that the mechanism driving SAM's improvements affect single-task and multi-task learning alike.

\subsection{When training data is limited}

We now switch gears and evaluate whether or not SAM helps when training data is scarce. Prior work~\citep{chen2021vision} showed that for vision models and tasks, SAM helps more when there is less training data to learn from. To test whether this holds for language, we do as follows: we subsample the training splits for both SuperGLUE and CBQA datasets at rates ranging from $2\%$ to $80\%$, and observe test performance when the public checkpoint is fine-tuned with and without SAM. SuperGLUE and CBQA results at a $5\%$ sampling rate are shown in Tables~\ref{tab:finetune_superglue_5} and \ref{tab:finetune_cbqa_5} respectively. In both cases we see again that SAM boosts performance across the board, adding, for example, a whopping $7.2\%$ relative improvement on the Base model on $5\%$ SuperGLUE and a relative $8.86\% / 16.6\%$ to F1/EM on Natural Questions.

Figure~\ref{fig:data-limited} plots the performance on the three CBQA tasks as a function of the sampling rate. We observe consistent gains from SAM across the size of the subsampled training set, with the relative improvement appearing largest when the subsampling rate is around $20\%$.

\begin{figure}[!t]
\centering
\includegraphics[width=\columnwidth]{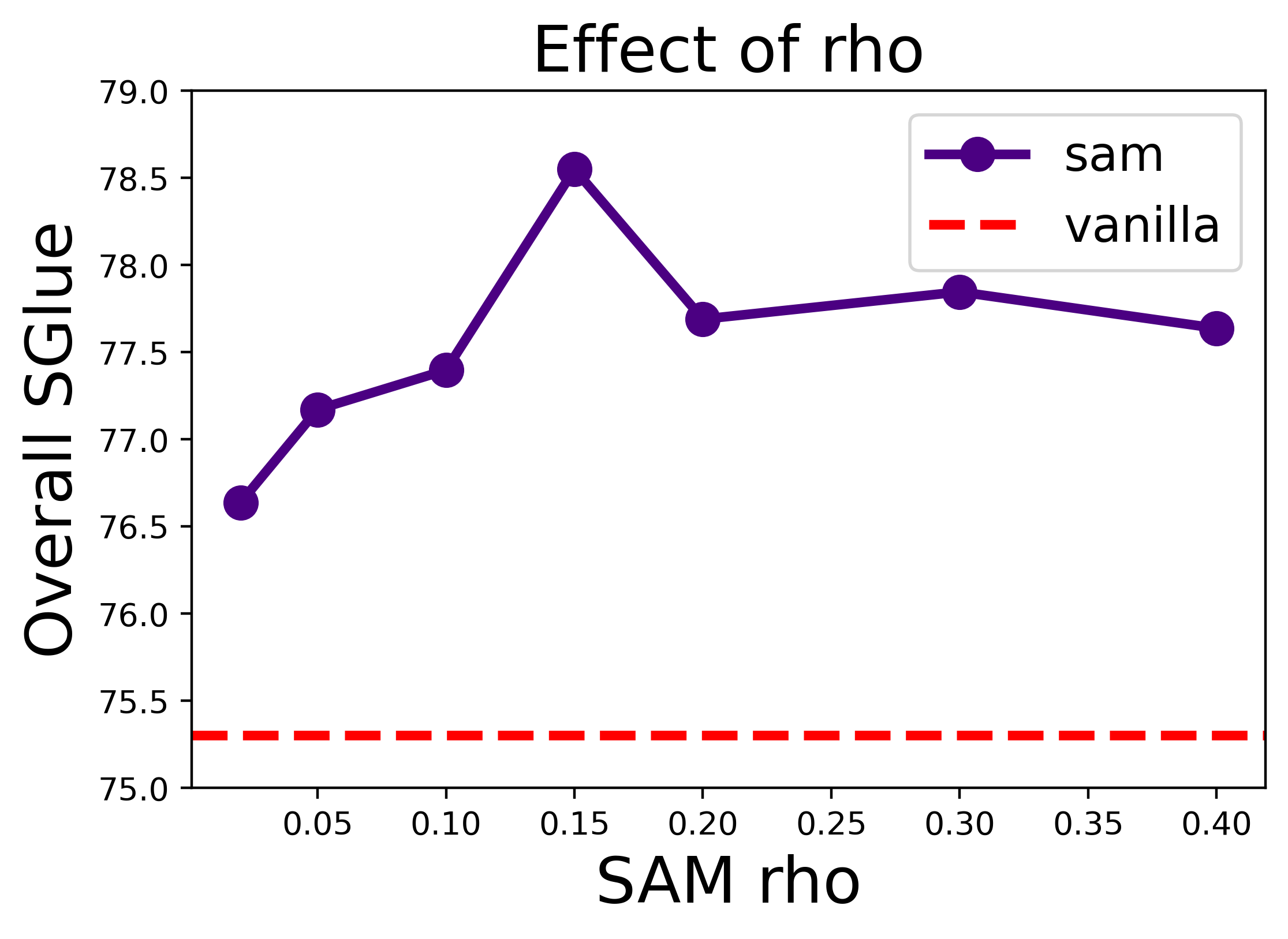} \\
\includegraphics[width=\columnwidth]{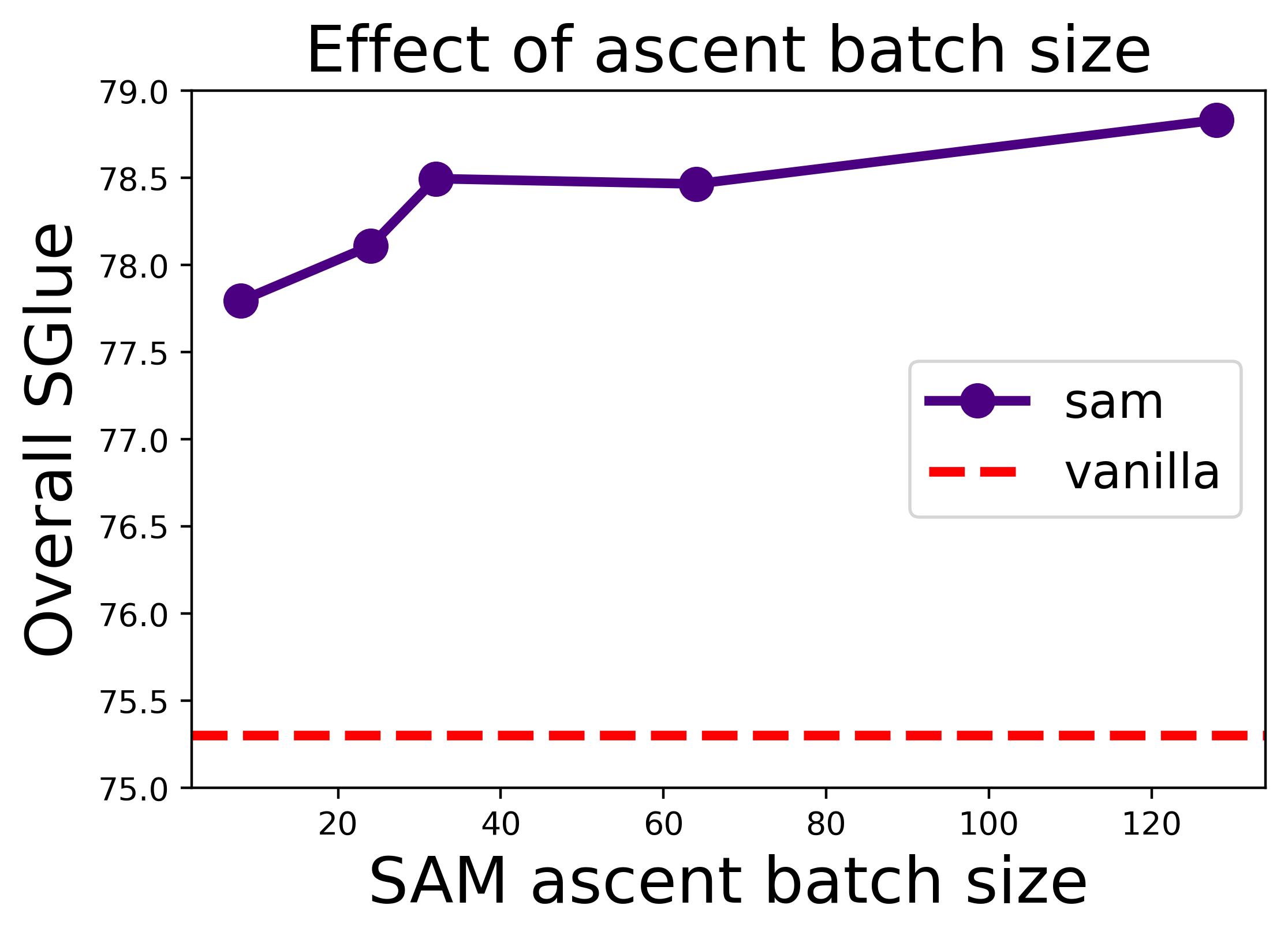} \\
\includegraphics[width=\columnwidth]{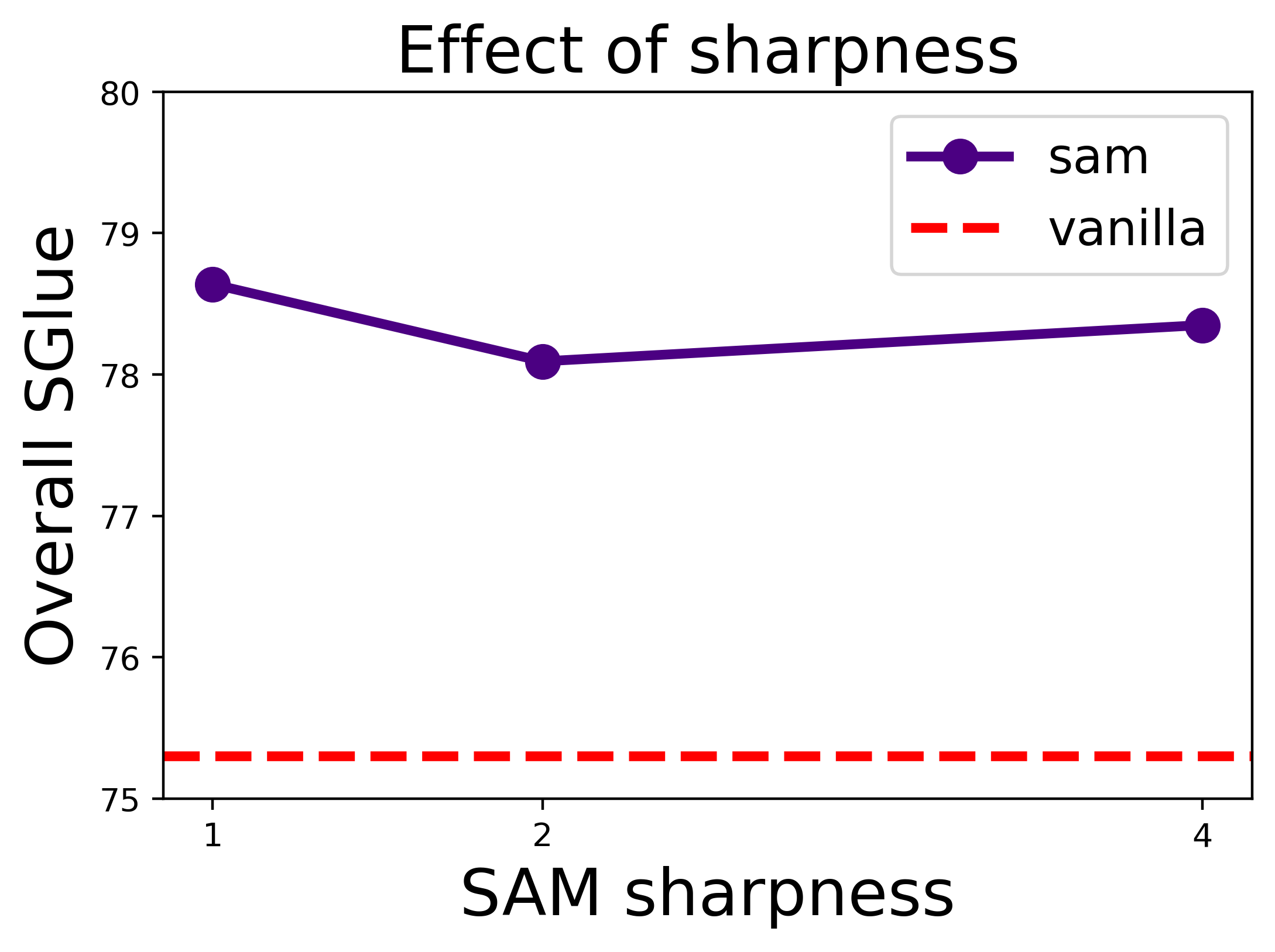} \\
\caption{Impact of SAM's hyper-parameters $\rho$, the ascent micro-batch size $a$, and the sharpness $m$ on the SuperGLUE benchmark for the Base model. Except for the ablated hyper-parameter, we use $\rho = 0.15$, $a = 32$, $m = 1$. For $\rho$, we see that all tested values perform better than fine-tuning without SAM. However, $0.15$ is a ``sweet spot,'' performing better than values below or above it. For the ascent micro-batch size $a$, we see that when the descent batch size is 128, there is improvement as $a$ is increased to 32 but little past this point. Thus, setting $a$ to be 1/4-th the descent batch size provides a good trade-off between performance and computational overhead. Increasing the sharpness $m$ does not help (in fact it hurts) here. Full results are shown in the Appendix.}
\label{fig:rho-ablation}
\end{figure}

\subsection{Sensitivity to hyper-parameters}
Figure~\ref{fig:rho-ablation} shows the impact of SAM's hyper-parameters $\rho$, the ascent micro-batch size $a$, and the sharpness factor $m$ on the (full) SuperGLUE benchmark for the Base model. For $\rho$, we see that all tested values perform better than fine-tuning without SAM. However, $0.15$ is a ``sweet spot,'' performing better than values below or above it. Thus, practitioners with little computational budget for hyper-parameter tuning may still see large gains by using a non-optimal $\rho$, while those with a generous budget should consider tuning. For the ascent micro-batch size $a$, we see that when the normal (descent) batch size is 128, there is improvement as $a$ is increased to 32 but little past this point. Thus, setting $a$ to be 1/4-th the descent batch size, as we do throughout our experiments, provides a good trade-off between performance and computational overhead. Increasing the sharpness $m$, where each of the $m$ ascent micro-batches has size $32/m$, does not improve performance here. We thus recommend a default of 1, which is the setting used across our experiments. Full results are shown in the Appendix.

\section{Conclusion}
To the best of our knowledge, this paper is the first to demonstrate how the recently-proposed Sharpness-Aware Minimization can be applied for fine-tuning the ubiquitous text-to-text Transformer (T5) and its multilingual counterpart mT5 on language tasks of broad interest. We thereby corroborate the already-documented success the method has had in the vision domain. Furthermore, we reveal SAM's benefits when data is limited by fine-tuning on subsamples of the original task training split. By approximating the ascent step of the algorithm via fewer samples, we show how large gains can be had across benchmarks and model sizes while adding only around $25\%$ additional compute and wall-clock training time. Our hope is that this work will spur SAM's adoption in the natural language processing community the way it is starting to in the vision one.

\bibliography{main}

\begin{thebibliography}{43}
\expandafter\ifx\csname natexlab\endcsname\relax\def\natexlab#1{#1}\fi

\bibitem[{Aghajanyan et~al.(2020)Aghajanyan, Shrivastava, Gupta, Goyal,
  Zettlemoyer, and Gupta}]{aghajanyan2020better}
Armen Aghajanyan, Akshat Shrivastava, Anchit Gupta, Naman Goyal, Luke
  Zettlemoyer, and Sonal Gupta. 2020.
\newblock Better fine-tuning by reducing representational collapse.
\newblock \emph{arXiv preprint arXiv:2008.03156}.

\bibitem[{Anil et~al.(2020)Anil, Gupta, Koren, Regan, and
  Singer}]{anil2020scalable}
Rohan Anil, Vineet Gupta, Tomer Koren, Kevin Regan, and Yoram Singer. 2020.
\newblock Scalable second order optimization for deep learning.
\newblock \emph{arXiv preprint arXiv:2002.09018}.

\bibitem[{Barrett and Dherin(2020)}]{barrett2020implicit}
David~GT Barrett and Benoit Dherin. 2020.
\newblock Implicit gradient regularization.
\newblock \emph{arXiv preprint arXiv:2009.11162}.

\bibitem[{Berant et~al.(2013)Berant, Chou, Frostig, and
  Liang}]{berant2013semantic}
Jonathan Berant, Andrew Chou, Roy Frostig, and Percy Liang. 2013.
\newblock Semantic parsing on freebase from question-answer pairs.
\newblock In \emph{Proceedings of the 2013 conference on empirical methods in
  natural language processing}, pages 1533--1544.

\bibitem[{Bradbury et~al.(2018)Bradbury, Frostig, Hawkins, Johnson, Leary,
  Maclaurin, Necula, Paszke, Vander{P}las, Wanderman-{M}ilne, and
  Zhang}]{jax2018github}
James Bradbury, Roy Frostig, Peter Hawkins, Matthew~James Johnson, Chris Leary,
  Dougal Maclaurin, George Necula, Adam Paszke, Jake Vander{P}las, Skye
  Wanderman-{M}ilne, and Qiao Zhang. 2018.
\newblock \href {http://github.com/google/jax} {{JAX}: composable
  transformations of {P}ython+{N}um{P}y programs}.

\bibitem[{Brock et~al.(2021)Brock, De, Smith, and Simonyan}]{brock2021high}
Andrew Brock, Soham De, Samuel~L Smith, and Karen Simonyan. 2021.
\newblock High-performance large-scale image recognition without normalization.
\newblock \emph{arXiv preprint arXiv:2102.06171}.

\bibitem[{Chaudhari et~al.(2019)Chaudhari, Choromanska, Soatto, LeCun,
  Baldassi, Borgs, Chayes, Sagun, and Zecchina}]{chaudhari2019entropy}
Pratik Chaudhari, Anna Choromanska, Stefano Soatto, Yann LeCun, Carlo Baldassi,
  Christian Borgs, Jennifer Chayes, Levent Sagun, and Riccardo Zecchina. 2019.
\newblock Entropy-sgd: Biasing gradient descent into wide valleys.
\newblock \emph{Journal of Statistical Mechanics: Theory and Experiment},
  2019(12):124018.

\bibitem[{Chen et~al.(2021)Chen, Hsieh, and Gong}]{chen2021vision}
Xiangning Chen, Cho-Jui Hsieh, and Boqing Gong. 2021.
\newblock When vision transformers outperform resnets without pretraining or
  strong data augmentations.
\newblock \emph{arXiv preprint arXiv:2106.01548}.

\bibitem[{Clark et~al.(2020)Clark, Choi, Collins, Garrette, Kwiatkowski,
  Nikolaev, and Palomaki}]{clark2020tydi}
Jonathan~H Clark, Eunsol Choi, Michael Collins, Dan Garrette, Tom Kwiatkowski,
  Vitaly Nikolaev, and Jennimaria Palomaki. 2020.
\newblock Tydi qa: A benchmark for information-seeking question answering in
  typologically diverse languages.
\newblock \emph{Transactions of the Association for Computational Linguistics},
  8:454--470.

\bibitem[{Devlin et~al.(2018)Devlin, Chang, Lee, and
  Toutanova}]{devlin2018bert}
Jacob Devlin, Ming-Wei Chang, Kenton Lee, and Kristina Toutanova. 2018.
\newblock Bert: Pre-training of deep bidirectional transformers for language
  understanding.
\newblock \emph{arXiv preprint arXiv:1810.04805}.

\bibitem[{Dosovitskiy et~al.(2020)Dosovitskiy, Beyer, Kolesnikov, Weissenborn,
  Zhai, Unterthiner, Dehghani, Minderer, Heigold, Gelly
  et~al.}]{dosovitskiy2020image}
Alexey Dosovitskiy, Lucas Beyer, Alexander Kolesnikov, Dirk Weissenborn,
  Xiaohua Zhai, Thomas Unterthiner, Mostafa Dehghani, Matthias Minderer, Georg
  Heigold, Sylvain Gelly, et~al. 2020.
\newblock An image is worth 16x16 words: Transformers for image recognition at
  scale.
\newblock \emph{arXiv preprint arXiv:2010.11929}.

\bibitem[{Foret et~al.(2020)Foret, Kleiner, Mobahi, and
  Neyshabur}]{foret2020sharpness}
Pierre Foret, Ariel Kleiner, Hossein Mobahi, and Behnam Neyshabur. 2020.
\newblock Sharpness-aware minimization for efficiently improving
  generalization.
\newblock \emph{arXiv preprint arXiv:2010.01412}.

\bibitem[{Gupta et~al.(2018)Gupta, Koren, and Singer}]{gupta2018shampoo}
Vineet Gupta, Tomer Koren, and Yoram Singer. 2018.
\newblock Shampoo: Preconditioned stochastic tensor optimization.
\newblock In \emph{International Conference on Machine Learning}, pages
  1842--1850. PMLR.

\bibitem[{He et~al.(2020)He, Liu, Gao, and Chen}]{he2020deberta}
Pengcheng He, Xiaodong Liu, Jianfeng Gao, and Weizhu Chen. 2020.
\newblock Deberta: Decoding-enhanced bert with disentangled attention.
\newblock \emph{arXiv preprint arXiv:2006.03654}.

\bibitem[{Hendrycks et~al.(2021)Hendrycks, Basart, Mu, Kadavath, Wang, Dorundo,
  Desai, Zhu, Parajuli, Guo et~al.}]{hendrycks2021many}
Dan Hendrycks, Steven Basart, Norman Mu, Saurav Kadavath, Frank Wang, Evan
  Dorundo, Rahul Desai, Tyler Zhu, Samyak Parajuli, Mike Guo, et~al. 2021.
\newblock The many faces of robustness: A critical analysis of
  out-of-distribution generalization.
\newblock In \emph{Proceedings of the IEEE/CVF International Conference on
  Computer Vision}, pages 8340--8349.

\bibitem[{Hendrycks and Dietterich(2019)}]{hendrycks2019benchmarking}
Dan Hendrycks and Thomas Dietterich. 2019.
\newblock Benchmarking neural network robustness to common corruptions and
  perturbations.
\newblock \emph{arXiv preprint arXiv:1903.12261}.

\bibitem[{Hinton et~al.(2015)Hinton, Vinyals, and Dean}]{hinton2015distilling}
Geoffrey Hinton, Oriol Vinyals, and Jeff Dean. 2015.
\newblock Distilling the knowledge in a neural network.
\newblock \emph{arXiv preprint arXiv:1503.02531}.

\bibitem[{Ioffe and Szegedy(2015)}]{ioffe2015batch}
Sergey Ioffe and Christian Szegedy. 2015.
\newblock Batch normalization: Accelerating deep network training by reducing
  internal covariate shift.
\newblock \emph{arXiv preprint arXiv:1502.03167}.

\bibitem[{Jiang et~al.(2019)Jiang, He, Chen, Liu, Gao, and
  Zhao}]{jiang2019smart}
Haoming Jiang, Pengcheng He, Weizhu Chen, Xiaodong Liu, Jianfeng Gao, and Tuo
  Zhao. 2019.
\newblock Smart: Robust and efficient fine-tuning for pre-trained natural
  language models through principled regularized optimization.
\newblock \emph{arXiv preprint arXiv:1911.03437}.

\bibitem[{Joshi et~al.(2017)Joshi, Choi, Weld, and
  Zettlemoyer}]{joshi2017triviaqa}
Mandar Joshi, Eunsol Choi, Daniel~S Weld, and Luke Zettlemoyer. 2017.
\newblock Triviaqa: A large scale distantly supervised challenge dataset for
  reading comprehension.
\newblock \emph{arXiv preprint arXiv:1705.03551}.

\bibitem[{Kleinberg et~al.(2018)Kleinberg, Li, and
  Yuan}]{kleinberg2018alternative}
Bobby Kleinberg, Yuanzhi Li, and Yang Yuan. 2018.
\newblock An alternative view: When does sgd escape local minima?
\newblock In \emph{International Conference on Machine Learning}, pages
  2698--2707. PMLR.

\bibitem[{Kwiatkowski et~al.(2019)Kwiatkowski, Palomaki, Redfield, Collins,
  Parikh, Alberti, Epstein, Polosukhin, Devlin, Lee
  et~al.}]{kwiatkowski2019natural}
Tom Kwiatkowski, Jennimaria Palomaki, Olivia Redfield, Michael Collins, Ankur
  Parikh, Chris Alberti, Danielle Epstein, Illia Polosukhin, Jacob Devlin,
  Kenton Lee, et~al. 2019.
\newblock Natural questions: a benchmark for question answering research.
\newblock \emph{Transactions of the Association for Computational Linguistics},
  7:453--466.

\bibitem[{Kwon et~al.(2021)Kwon, Kim, Park, and Choi}]{kwon2021asam}
Jungmin Kwon, Jeongseop Kim, Hyunseo Park, and In~Kwon Choi. 2021.
\newblock Asam: Adaptive sharpness-aware minimization for scale-invariant
  learning of deep neural networks.
\newblock \emph{arXiv preprint arXiv:2102.11600}.

\bibitem[{MacKay(1995)}]{mackay1995probable}
David~JC MacKay. 1995.
\newblock Probable networks and plausible predictions-a review of practical
  bayesian methods for supervised neural networks.
\newblock \emph{Network: computation in neural systems}, 6(3):469.

\bibitem[{Martens and Grosse(2015)}]{martens2015optimizing}
James Martens and Roger Grosse. 2015.
\newblock Optimizing neural networks with kronecker-factored approximate
  curvature.
\newblock In \emph{International conference on machine learning}, pages
  2408--2417. PMLR.

\bibitem[{Mobahi et~al.(2020)Mobahi, Farajtabar, and Bartlett}]{mobahi2020self}
Hossein Mobahi, Mehrdad Farajtabar, and Peter~L Bartlett. 2020.
\newblock Self-distillation amplifies regularization in hilbert space.
\newblock \emph{arXiv preprint arXiv:2002.05715}.

\bibitem[{M{\"u}ller et~al.(2019)M{\"u}ller, Kornblith, and
  Hinton}]{muller2019does}
Rafael M{\"u}ller, Simon Kornblith, and Geoffrey Hinton. 2019.
\newblock When does label smoothing help?
\newblock \emph{arXiv preprint arXiv:1906.02629}.

\bibitem[{Radford et~al.(2019)Radford, Wu, Child, Luan, Amodei, Sutskever
  et~al.}]{radford2019language}
Alec Radford, Jeffrey Wu, Rewon Child, David Luan, Dario Amodei, Ilya
  Sutskever, et~al. 2019.
\newblock Language models are unsupervised multitask learners.
\newblock \emph{OpenAI blog}, 1(8):9.

\bibitem[{Raffel et~al.(2019)Raffel, Shazeer, Roberts, Lee, Narang, Matena,
  Zhou, Li, and Liu}]{raffel2019exploring}
Colin Raffel, Noam Shazeer, Adam Roberts, Katherine Lee, Sharan Narang, Michael
  Matena, Yanqi Zhou, Wei Li, and Peter~J Liu. 2019.
\newblock Exploring the limits of transfer learning with a unified text-to-text
  transformer.
\newblock \emph{arXiv preprint arXiv:1910.10683}.

\bibitem[{Roberts et~al.(2020)Roberts, Raffel, and Shazeer}]{roberts2020much}
Adam Roberts, Colin Raffel, and Noam Shazeer. 2020.
\newblock How much knowledge can you pack into the parameters of a language
  model?
\newblock \emph{arXiv preprint arXiv:2002.08910}.

\bibitem[{Shazeer and Stern(2018)}]{shazeer2018adafactor}
Noam Shazeer and Mitchell Stern. 2018.
\newblock Adafactor: Adaptive learning rates with sublinear memory cost.
\newblock In \emph{International Conference on Machine Learning}, pages
  4596--4604. PMLR.

\bibitem[{{Shirish Keskar} et~al.(2016){Shirish Keskar}, {Mudigere}, {Nocedal},
  {Smelyanskiy}, and {Tang}}]{keshar}
Nitish {Shirish Keskar}, Dheevatsa {Mudigere}, Jorge {Nocedal}, Mikhail
  {Smelyanskiy}, and Ping Tak~Peter {Tang}. 2016.
\newblock \href {http://arxiv.org/abs/1609.04836} {{On Large-Batch Training for
  Deep Learning: Generalization Gap and Sharp Minima}}.
\newblock \emph{arXiv e-prints}, page arXiv:1609.04836.

\bibitem[{Smith et~al.(2021)Smith, Dherin, Barrett, and De}]{smith2021origin}
Samuel~L Smith, Benoit Dherin, David~GT Barrett, and Soham De. 2021.
\newblock On the origin of implicit regularization in stochastic gradient
  descent.
\newblock \emph{arXiv preprint arXiv:2101.12176}.

\bibitem[{Smith and Le(2017)}]{smith2017bayesian}
Samuel~L Smith and Quoc~V Le. 2017.
\newblock A bayesian perspective on generalization and stochastic gradient
  descent.
\newblock \emph{arXiv preprint arXiv:1710.06451}.

\bibitem[{Srivastava et~al.(2014)Srivastava, Hinton, Krizhevsky, Sutskever, and
  Salakhutdinov}]{srivastava2014dropout}
Nitish Srivastava, Geoffrey Hinton, Alex Krizhevsky, Ilya Sutskever, and Ruslan
  Salakhutdinov. 2014.
\newblock Dropout: a simple way to prevent neural networks from overfitting.
\newblock \emph{The journal of machine learning research}, 15(1):1929--1958.

\bibitem[{Tolstikhin et~al.(2021)Tolstikhin, Houlsby, Kolesnikov, Beyer, Zhai,
  Unterthiner, Yung, Keysers, Uszkoreit, Lucic et~al.}]{tolstikhin2021mlp}
Ilya Tolstikhin, Neil Houlsby, Alexander Kolesnikov, Lucas Beyer, Xiaohua Zhai,
  Thomas Unterthiner, Jessica Yung, Daniel Keysers, Jakob Uszkoreit, Mario
  Lucic, et~al. 2021.
\newblock Mlp-mixer: An all-mlp architecture for vision.
\newblock \emph{arXiv preprint arXiv:2105.01601}.

\bibitem[{Vaswani et~al.(2017)Vaswani, Shazeer, Parmar, Uszkoreit, Jones,
  Gomez, Kaiser, and Polosukhin}]{vaswani2017attention}
Ashish Vaswani, Noam Shazeer, Niki Parmar, Jakob Uszkoreit, Llion Jones,
  Aidan~N Gomez, {\L}ukasz Kaiser, and Illia Polosukhin. 2017.
\newblock Attention is all you need.
\newblock In \emph{Advances in neural information processing systems}, pages
  5998--6008.

\bibitem[{Wang et~al.(2019)Wang, Pruksachatkun, Nangia, Singh, Michael, Hill,
  Levy, and Bowman}]{wang2019superglue}
Alex Wang, Yada Pruksachatkun, Nikita Nangia, Amanpreet Singh, Julian Michael,
  Felix Hill, Omer Levy, and Samuel~R Bowman. 2019.
\newblock Superglue: A stickier benchmark for general-purpose language
  understanding systems.
\newblock \emph{arXiv preprint arXiv:1905.00537}.

\bibitem[{Wang et~al.(2018)Wang, Singh, Michael, Hill, Levy, and
  Bowman}]{wang2018glue}
Alex Wang, Amanpreet Singh, Julian Michael, Felix Hill, Omer Levy, and Samuel~R
  Bowman. 2018.
\newblock Glue: A multi-task benchmark and analysis platform for natural
  language understanding.
\newblock \emph{arXiv preprint arXiv:1804.07461}.

\bibitem[{Wilson and Izmailov(2020)}]{wilson2020bayesian}
Andrew~Gordon Wilson and Pavel Izmailov. 2020.
\newblock Bayesian deep learning and a probabilistic perspective of
  generalization.
\newblock \emph{arXiv preprint arXiv:2002.08791}.

\bibitem[{Xue et~al.(2020)Xue, Constant, Roberts, Kale, Al-Rfou, Siddhant,
  Barua, and Raffel}]{xue2020mt5}
Linting Xue, Noah Constant, Adam Roberts, Mihir Kale, Rami Al-Rfou, Aditya
  Siddhant, Aditya Barua, and Colin Raffel. 2020.
\newblock mt5: A massively multilingual pre-trained text-to-text transformer.
\newblock \emph{arXiv preprint arXiv:2010.11934}.

\bibitem[{Zhang et~al.(2017)Zhang, Cisse, Dauphin, and
  Lopez-Paz}]{zhang2017mixup}
Hongyi Zhang, Moustapha Cisse, Yann~N Dauphin, and David Lopez-Paz. 2017.
\newblock mixup: Beyond empirical risk minimization.
\newblock \emph{arXiv preprint arXiv:1710.09412}.

\bibitem[{Zhu et~al.(2019)Zhu, Cheng, Gan, Sun, Goldstein, and
  Liu}]{zhu2019freelb}
Chen Zhu, Yu~Cheng, Zhe Gan, Siqi Sun, Tom Goldstein, and Jingjing Liu. 2019.
\newblock Freelb: Enhanced adversarial training for natural language
  understanding.
\newblock \emph{arXiv preprint arXiv:1909.11764}.

\end{thebibliography}
\bibliographystyle{acl_natbib}

\clearpage

\section{Appendix}

\subsection{TyDiQA-GoldP Results}
Table~\ref{tab:tydiqa_full} shows the per-language TyDiQA-GoldP scores. We found that the multilingual mT5 model benefited from a smaller $\rho$ than the vanilla T5 model.

\subsection{Impact of hyper-parameters on SuperGLUE}
Table~\ref{tab:superglue_ablation_full} shows the full (no subsampling) SuperGLUE results for the Base model for different hyper-parameter choices.

\begin{table*}[!t]
\tiny
    \centering
    \begin{tabular}{l|c|ccccccccc}
    \toprule
        Model &	avg (F1/EM) & en & ar & bn & fi & id & ko & ru & sw & te \\
\midrule
Small & 73.4 / 62.1 & 66.3 / 54.5 & 78.0 / 62.8 & 69.4 / 60.2 & 73.7 / 60.9 & 77.8 / 65.0 & 64.1 / 55.8 & 69.5 / 56.7 & 78.0 / 68.9 & 84.0 / 74.4 \\
Small + SAM (0.02) & 74.3 / 63.0 & 67.1 / 55.7 & 79.2 / 64.9 & 68.4 / 57.5 & 75.0 / 61.8 & 78.8 / 67.6 & 64.0 / 54.3 & 72.0 / 58.0 & 79.9 / 72.1 & 84.2 / 75.0 \\
\midrule
Base & 81.6 / 71.0 & 76.7 / 65.7 & 84.2 / 70.4 & 81.7 / 71.7 & 81.6 / 69.3 & 85.1 / 74.2 & 73.6 / 66.3 & 78.8 / 64.7 & 84.4 / 77.2 & 87.9 / 79.4 \\
Base + SAM (0.02) & 82.0 / 71.3 & 76.1 / 65.0 & 83.9 / 70.2 & 84.4 / 75.2 & 81.1 / 69.6 & 85.4 / 74.0 & 74.5 / 66.7 & 79.5 / 65.3 & 84.8 / 76.4 & 88.8 / 79.1 \\
\midrule
Large & 85.6 / 75.3 & 81.2 / 70.0 & 86.6 / 73.1 & 86.3 / 77.9 & 84.6 / 71.6 & 87.7 / 77.9 & 81.1 / 72.1 & 84.0 / 72.8 & 88.7 / 81.2 & 90.3 / 81.5 \\
Large + SAM (0.02) & 85.9 / 76.1 & 82.3 / 71.6 & 87.3 / 74.4 & 86.6 / 79.6 & 84.5 / 72.5 & 88.1 / 80.0 & 81.3 / 73.2 & 84.2 / 71.9 & 88.3 / 79.8 & 90.9 / 82.1 \\
\midrule
XL & 87.0 / 77.4 & 82.1 / 72.0 & 87.3 / 74.4 & 88.9 / 82.3 & 85.8 / 74.6 & 89.7 / 80.4 & 81.7 / 73.6 & 85.1 / 73.5 & 90.7 / 83.2 & 91.5 / 82.4 \\
XL + SAM (0.05) & 87.3 / 77.7 & 82.8 / 73.4 & 87.6 / 75.0 & 89.2 / 81.4 & 86.4 / 74.2 & 89.6 / 80.2 & 82.6 / 75.0 & 85.5 / 74.8 & 90.7 / 83.0 & 91.4 / 82.5 \\
        \bottomrule
    \end{tabular}
    \caption{Full Results for TyDiQA-GoldP.}
    \label{tab:tydiqa_full}
\end{table*}

\begin{table*}[!t]
\small
    \centering
    \begin{tabular}{l|c|ccccccccc}
    \toprule
        Model &	SGlue &	BoolQ &	CB & 	CoPA & MultiRC &	ReCoRD &	RTE & 	WiC &	WSC \\ 
        \midrule
Base + SAM (0.02) & 76.6 & 80.5 & 92.4 / 92.9 & 73.0 & 76.2 / 36.8 & 77.3 / 76.4 & 81.9 & 70.8 & 80.8 \\
Base + SAM (0.05) & 77.2 & 80.3 & 97.4 / 96.4 & 73.0 & 76.4 / 37.9 & 77.8 / 76.9 & 83.8 & 71.9 & 76.9 \\
Base + SAM (0.1) & 77.4 & 81.7 & 94.8 / 94.6 & 72.0 & 76.8 / 38.3 & 79.3 / 78.3 & 83.8 & 72.7 & 77.9 \\
Base + SAM (0.15) & 78.5 & 82.2 & 93.7 / 94.6 & 78.0 & 77.5 / 39.1 & 78.2 / 77.2 & 85.9 & 70.4 & 81.7 \\
Base + SAM (0.2) & 77.7 & 82.3 & 95.0 / 96.4 & 75.0 & 77.2 / 39.5 & 77.9 / 76.8 & 84.1 & 71.8 & 76.9 \\
Base + SAM (0.3) & 77.8 & 81.1 & 93.6 / 94.6 & 74.0 & 77.4 / 40.0 & 79.8 / 78.7 & 85.9 & 70.8 & 78.8 \\
Base + SAM (0.4) & 77.6 & 80.6 & 94.3 / 96.4 & 79.0 & 76.3 / 37.5 & 77.9 / 76.9 & 81.9 & 71.0 & 78.8 \\
\midrule
\midrule
Base + SAM (8) & 77.8 & 81.9 & 97.4 / 96.4 & 72.0 & 77.5 / 40.8 & 78.6 / 77.5 & 84.5 & 71.0 & 78.8 \\
Base + SAM (24) & 78.1 & 82.5 & 97.4 / 96.4 & 74.0 & 77.2 / 40.2 & 79.2 / 78.4 & 83.8 & 69.4 & 80.8 \\
Base + SAM (32) & 78.5 & 81.9 & 96.1 / 94.6 & 77.0 & 77.6 / 40.3 & 78.9 / 77.8 & 85.6 & 71.9 & 78.8 \\
Base + SAM (64) & 78.5 & 82.3 & 96.1 / 94.6 & 79.0 & 78.4 / 40.8 & 79.3 / 78.2 & 84.5 & 71.3 & 76.9 \\
Base + SAM (128) & 78.8 & 82.4 & 97.4 / 96.4 & 74.0 & 77.6 / 40.6 & 79.0 / 78.0 & 87.0 & 72.9 & 79.8 \\
\midrule
\midrule
Base + SAM (1) & 78.6 & 81.5 & 96.1 / 96.4 & 73.0 & 77.2 / 40.0 & 79.0 / 77.9 & 87.0 & 72.6 & 81.7 \\
Base + SAM (2) & 78.1 & 81.8 & 96.1 / 94.6 & 77.0 & 76.5 / 37.6 & 78.5 / 77.6 & 84.8 & 71.8 & 78.8 \\
Base + SAM (4) & 78.3 & 82.3 & 97.4 / 96.4 & 75.0 & 77.3 / 38.5 & 78.2 / 77.1 & 85.2 & 72.1 & 79.8 \\
\midrule
\bottomrule
    \end{tabular}
    \caption{SuperGLUE results for Base for different values of $\rho$ (\textbf{top}), ascent micro-batch size $a$ (\textbf{middle}), and sharpness $m$ (\textbf{bottom}). Except for the ablated hyper-parameter, we use $\rho = 0.15$, $a = 32$, $m = 1$. }
    \label{tab:superglue_ablation_full}
\end{table*}

\end{document}